\documentclass[10pt]{article} % For LaTeX2e

\usepackage[preprint]{rlj} % Should be uncommented for preprint versions

\usepackage{amssymb}            % Defines common symbols like \mathbb R
\usepackage{mathtools}          % Extends amsmath, providing common math tools
\usepackage{mathrsfs}           % Enables \mathscr, which can work in cases that \mathcal does not
\usepackage{graphicx}           % For including images
\usepackage{subcaption}         % Allows for the use of subfigures and subcaptions
\usepackage[space]{grffile}     % For spaces in image names
\usepackage{url}                % For displaying URLs
\usepackage{lipsum}             % For placeholder text
\usepackage{glossaries}
\newacronym{rl}{RL}{Reinforcement Learning}
\newacronym{map}{MAP}{Model and Acceleration-Based Pursuit}
\newacronym{pp}{PP}{Pure Pursuit}
\newacronym{sac}{SAC}{Soft Actor-Critic}
\newacronym{ftg}{FTG}{Follow the Gap}
\newacronym{her}{HER}{Hindsight Experience Replay}
\newacronym{mpc}{MPC}{Model Predictive Control}
\newacronym{td}{TD}{Temporal-Difference}
\newacronym{f1tenth}{F1TENTH}{F1 Tenth Racing Platform}
\newacronym{vesc}{VESC}{Vedder Electronic Speed Controller}
\newacronym{sd}{SD}{Standard Deviation}
\newacronym{tpu}{TPU}{Thermoplastic Polyurethane}
\newacronym{api}{API}{Application Programming Interface}
\newacronym{e2e}{E2E}{End-to-End}
\newacronym{sota}{SotA}{State-of-the-Art}
\newacronym{ros}{ROS}{Robot Operating System}
\newacronym{imu}{IMU}{Inertial Measurement Unit}
\newacronym{ar}{AR}{Autonomous Racing}
\newacronym{hdra}{HDRA}{Heuristic Delayed Reward Adjustment}
\newacronym{dqn}{DQN}{Deep Q Network}
\newacronym{td3}{TD3}{Twin-Delayed Deep Deterministic}
\newacronym{slam}{SLAM}{ Simultaneous Localization and Mapping}
\newacronym{s2r}{sim-to-real}{Simulation-to-Reality}
\newacronym{lidar}{LiDAR}{Light Detection And Ranging}
\newacronym{gpu}{GPU}{Graphical Processing Unit}
\newacronym{yolo}{YOLO}{You Only Look Once}
\newacronym{ssd}{SSD}{Single-Shot Detector}
\newacronym{mbpo}{MBPO}{Model Based Policy Optimization}

\usepackage{booktabs}
\usepackage{algorithm}
\usepackage{algpseudocode}
\usepackage{wrapfig}
\usepackage{multirow}
\usepackage[british]{babel}
\usepackage{cleveref}
\usepackage{siunitx} % for si units
\usepackage{bm} % for a more powerful math boldification

\title{Drive Fast, Learn Faster: On-Board RL for High~Performance~Autonomous~Racing}
\setrunningtitle{On-Board Reinforcement Learning}

\author{Benedict Hildisch\textsuperscript{1,$\dagger$}, Edoardo Ghignone\textsuperscript{1,$\dagger$}, Nicolas Baumann\textsuperscript{1}, Cheng Hu\textsuperscript{2}, Andrea Carron\textsuperscript{3}, Michele Magno\textsuperscript{1}}

\emails{edoardo.ghignone@pbl.ee.ethz.ch}

\affiliations{
$^{1}$\textbf{Center for Project-Based Learning, D-ITET, ETH Zurich}\\
$^{2}$\textbf{Department of Control Science and Engineering, Zhejiang University}\\
$^{3}$\textbf{Institute for Dynamic Systems and Control, D-MAVT, ETH Zurich}\\
\par % If including additional comments like below, use \par to add some whitespace. 
$^\dagger$ Equal Contribution
}

\contribution{
    This paper presents an efficient \gls{rl} architecture, based on the residual policy structure, that enables uninterrupted on-board learning at an unprecedented level of performance. The architecture is compared to an \gls{e2e} structure, showing how the residual policy greatly increases efficiency by simplifying the control task.    
    }
    {
    The work in \citep{fastrlap} presented a high-speed driving solution with on-vehicle data only but failed to compare with \gls{sota} traditional methods for racing, exhibiting a performance gap.
    We show how a residual structure, previously used by \cite{trumpp2023residual} in simulation, can be used to efficiently train in reality only and overtake classical \gls{sota} performance.
    }

\contribution{
    This paper compares different baseline controllers with the residual \gls{rl} policy, showing how the proposed architecture is easily adaptable and improves on all the methods. 
    The fastest agent in the comparison eventually turns out to be the fastest F1TENTH algorithm when compared to previous best results.
    }
    {
    The work in \citep{trumpp2023residual} only used \gls{pp} as a baseline controller, while we show that the residual architecture can also adapt to different ones.     
    }

\contribution{
    This paper provides an ablation study in the F1TENTH simulation environment of the different \gls{rl} architectural choices, showing how the different parts affect efficiency and training consistency. 
    }
    {
    Some of the architectural choices were motivated by previous ablation studies (multi-step \gls{td} by \cite{barth-maron2018distributional}, asynchronous training by \cite{async}). Here the full combination is tested in the \gls{ar} environment. Furthermore, the introduction of the novel \gls{hdra} is analyzed.
    }

\contribution{
    This paper provides a generalization study showing how the residual policy behaves with a different type of track and tires in a zero-shot and few-shot setup.
    }
    {
    None
    }

 % Your keywords

\summary{
Autonomous racing presents unique challenges due to its non-linear dynamics, the high speed involved, and the critical need for real-time decision-making under dynamic and unpredictable conditions.
Most traditional \gls{rl} approaches rely on extensive simulation-based pre-training, which faces crucial challenges in transfer effectively to real-world environments.
This paper introduces a robust on-board \gls{rl} framework for autonomous racing, designed to eliminate the dependency on simulation-based pre-training enabling direct real-world adaptation.
The proposed system introduces a refined \gls{sac} algorithm, leveraging a residual \gls{rl} structure to enhance classical controllers in real-time by integrating multi-step \gls{td} learning, an asynchronous training pipeline, and \gls{hdra} to improve sample efficiency and training stability.
The framework is validated through extensive experiments on the F1TENTH racing platform, where the residual \gls{rl} controller consistently outperforms the baseline controllers and achieves up to an 11.5\,\% reduction in lap times compared to the \gls{sota}
with only \qty{20}{\minute} of training.
Additionally, an \gls{e2e} \gls{rl} controller trained without a baseline controller surpasses the previous best results with sustained on-track learning. 
}

\begin{document}

\makeCover  % Create the cover page
\maketitle  % Make the title section

\glsresetall % to reset glossary after cover page
\begin{abstract}
Autonomous racing presents unique challenges due to its non-linear dynamics, the high speed involved, and the critical need for real-time decision-making under dynamic and unpredictable conditions.
Most traditional \gls{rl} approaches rely on extensive simulation-based pre-training, which faces crucial challenges in transfer effectively to real-world environments.
This paper introduces a robust on-board \gls{rl} framework for autonomous racing, designed to eliminate the dependency on simulation-based pre-training enabling direct real-world adaptation.
% By enabling direct training of \gls{rl} agents on physical racing platforms, the proposed approach removes the \acrlong{s2r} transfer challenges and enhances real-world adaptability.
% The framework integrates a residual \gls{rl} controller with classical controllers, providing fine-grained, real-time adjustments to optimize lap performance. The method incorporates in a modified \gls{sac} algorithm the following: multi-step temporal-difference learning, heuristic delayed reward adjustments, and an asynchronous training architecture to enhance sample efficiency and stabilize learning dynamics. 
The proposed system introduces a refined \gls{sac} algorithm, leveraging a residual \gls{rl} structure to enhance classical controllers in real-time by integrating multi-step \gls{td} learning, an asynchronous training pipeline, and \gls{hdra} to improve sample efficiency and training stability.
The framework is validated through extensive experiments on the F1TENTH racing platform, where the residual \gls{rl} controller consistently outperforms the baseline controllers and achieves up to an 11.5\,\% reduction in lap times compared to the \gls{sota}
with only \qty{20}{\minute} of training.
Additionally, an \gls{e2e} \gls{rl} controller trained without a baseline controller surpasses the previous best results with sustained on-track learning. 
These findings position the framework as a robust solution for high-performance autonomous racing and a promising direction for other real-time, dynamic autonomous systems.
\end{abstract}

%%%%%%%%%%%%%%%%%%%%%%%%%%%%%%%%%%%%%%%%%%%%%%%%%%%%%%%%%%%%%%%%
%% Section: Introduction
%%%%%%%%%%%%%%%%%%%%%%%%%%%%%%%%%%%%%%%%%%%%%%%%%%%%%%%%%%%%%%%%
\glsresetall
\section{Introduction}
\label{sec:introduction}

\gls{rl} has become ubiquitous in robotics, particularly on quadrupedal robots, \gls{rl} is now almost indispensable for locomotion, as it implicitly learns efficient heuristics for complex movements \cite{hoeller2024anymalparkour}.
Similarly, in autonomous drone racing, \gls{rl} has exhibited superhuman control capabilities, outperforming traditional approaches and human champions \cite{kaufmann2023champion}. These successes highlight the necessity of \gls{rl} in robotics, particularly for pushing robotic systems to their performance limits.

Despite its success in various robotic domains, \gls{rl} has not yet achieved widespread adoption in \gls{ar}. This is particularly notable, given that  \gls{ar} also involves agile control of non-linear systems posing significant challenges for robotic platforms \citep{betz2023tumfullstack}. Moreover, the few solutions found on \gls{ar} are showing impressive results only in simulation \citep{Fuchs2021, superhuman_rl}, while evaluations on physical race cars remain limited. One possible explanation is due to the fact that when deployed on real vehicles, \gls{rl}-based controllers are consistently slower than \gls{sota} classical \gls{ar} controllers \citep{brunnbauer2022latent, ghignone2025rlpp}. This suggests that the \gls{s2r} gap in \gls{ar} is particularly challenging to bridge.

One approach to mitigating this gap involves developing high-fidelity simulators that accurately model vehicle dynamics, adapting them to every possible physical environment.
While this approach has been effective, especially in tandem with data-driven techniques that improve simulation accuracy \citep{kaufmann2023champion}, it remains limited by inevitable mismatches between simulated and real-world physics and requires accurate system identification of the physical system and calibration of the simulator \citep{song2023MPCvsRL}. Instead, we take a different approach, bypassing these limitations entirely by training directly on the physical car and allowing the agent to learn and adapt to real-world dynamics from the start.
% However, tire dynamics are highly sensitive to variations in surface properties, temperature, and aerodynamic effects. \todo{rewrite this scientifically}
% Accurately simulating these interactions for all conditions would be computationally intractable.

\begin{figure}[ht]
    \begin{center}
        \includegraphics[width=\textwidth, trim={0 0 0 1.25cm}, clip]{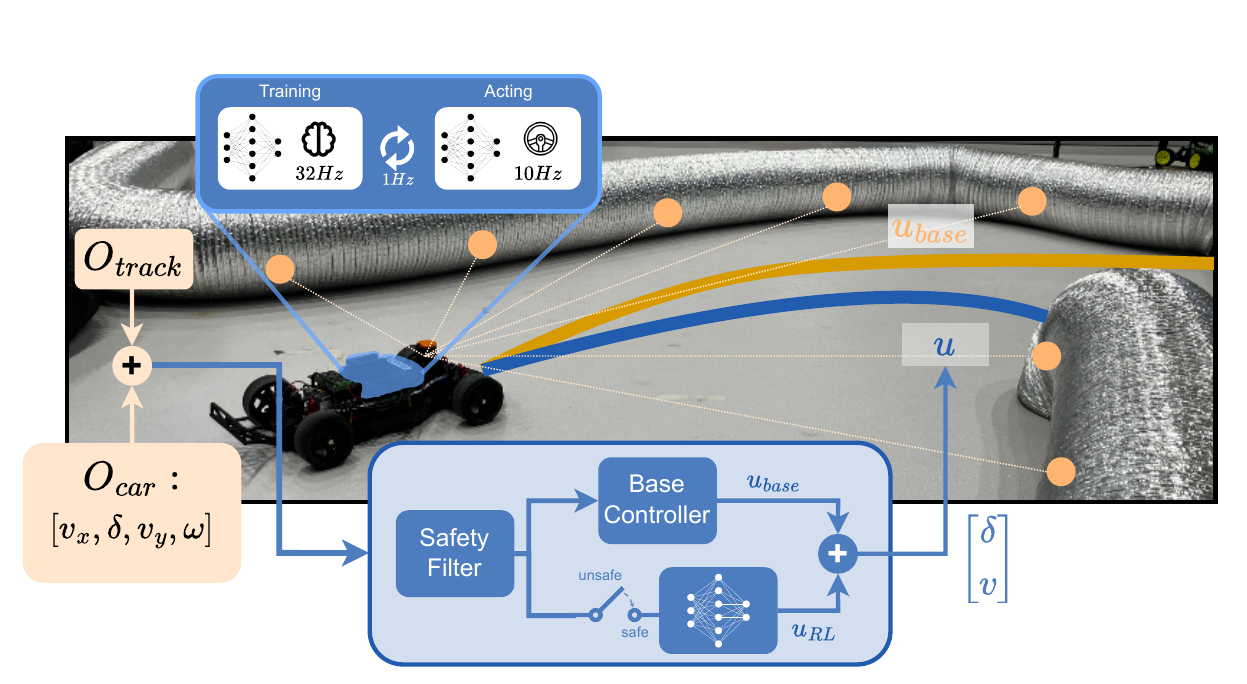}
    \end{center}
    \vspace{-0.5 cm}
    % \caption{Overview of the proposed \gls{rl} architecture, which processes observations $O_{car}$ --- providing proprioceptive information such as position and velocity --- and $O_{track}$ --- describing track boundaries and the reference trajectory (see \Cref{sec:observation} for more details) --- to generate control commands $u$ which are combined in a residual fashion from a conventional controller $u_{base}$ and a \gls{rl} agent. The \gls{rl} agent is trained directly on the racing car, mitigating the \gls{s2r} gap. Since the agent starts untrained and may encounter unsafe states, a safety filter is employed to guide the robot back to a safe state. Training happens entirely on-board in an asynchronous fashion: the \emph{Training} node samples and updates the policy at \qty{32}{\hertz}, the \emph{Acting} node provides inputs to the vehicle at \qty{10}{\hertz}, and the two policies are synchronized every second.}
    \caption{Overview of the proposed \gls{rl} architecture, which processes observations $O_{car}$, providing proprioceptive information such as position and velocity $(v_x,\,v_y)$, and $O_{track}$, describing track boundaries and a given reference trajectory to generate a residual control command $u_{RL}$. Such a command is additively combined with the output of a conventional controller $u_{base}$. Since the agent starts untrained and may encounter unsafe states, a safety filter is employed to guide the robot back to a safe state. Training happens entirely on-board in an asynchronous fashion: the \emph{Acting} node provides residual control commands at \qty{10}{\hertz}, while the training node updates the control policy at \qty{32}{\hertz}. The policy of the acting node is updated with the policy of the training node every second.}
    \label{fig:architecture}
\end{figure}

This work proposes an onboard \gls{rl}-based high-performance \gls{ar} controller that significantly and consistently outperforms \gls{sota} classical controllers in terms of lap time.
This work eliminates the \gls{s2r} gap by training the \gls{rl} agent  \citep{daydreamer, fastrlap} directly on the physical robot,  namely a 1:10 scaled F1TENTH race car \citep{f1tenth}, and validates the approach across multiple track layouts and friction conditions. 
With accurate real-field measurements the paper demonstrates that an \gls{rl} agent can surpass a high-performance, model-based \gls{sota} racing controller in real-world \gls{ar}, achieving up to a \qty{11.5}{\percent} % t mean rlmap vs t mean map on C track 
improvement in lap time, and enabling the physical robot to operate without speed constraints and over \qty{6.5}{\metre\per\second}. 
Moreover one important finding out of this approach is that direct on-robot training introduces a new challenge: unlike simulation, where massive parallelization can be leveraged to accelerate learning \citep{rudin2021learning}, real-world training remains inherently limited by the time needed for data collection.
To address this problem, a residual \gls{rl} structure is employed \citep{johannink2019residual}, enabling a fully untrained agent to learn to exceed the performance of the \gls{sota} controller within \qty{20}{\minute}. This paper further shows that such architecture significantly accelerates the training process, as an \gls{e2e} model (not residual), achieves similar but slower lap times in three times the amount of training.

The proposed training framework is based on the \gls{sac} algorithm and incorporates several modifications to facilitate on-board learning. Multi-step \gls{td} updates are utilized to manage delayed rewards \citep{sutton1998introduction}, while \gls{hdra}, inspired by \gls{her}, is employed to emphasize undesirable behaviors. Additionally, an asynchronous training pipeline is implemented, decoupling data collection from policy optimization, following the approach of \cite{async}. Finally, a safety filter allows for uninterrupted data collection, by gating potentially unsafe inputs and rerouting the vehicle to a reference safe trajectory. 
An overview of the proposed architecture is available in \Cref{fig:architecture}.

%%%%%%%%%%%%%%%%%%%%%%%%%%%%%%%%%%%%%%%%%%%%%%%%%%%%%%%%%%%%%%%%
%% Section: Related work
%%%%%%%%%%%%%%%%%%%%%%%%%%%%%%%%%%%%%%%%%%%%%%%%%%%%%%%%%%%%%%%%
\section{Related Work}
\label{sec:related_work}
 \gls{rl} has seen growing applications in \gls{ar} due to its potential to handle high-speed decision-making and complex, non-linear vehicle dynamics \citep{Fuchs2021, superhuman_rl}. While this section focuses on previous work that directly informs our approach, \Cref{appendix:related_works} gives a broader overview of the advancements in autonomous driving and its subfield \gls{ar}.

The \textbf{\Gls{s2r} gap} has been a persistent issue in applying \gls{rl} to physical \gls{ar} systems \citep{sim2real}. This gap stems from the inability of simulators to accurately replicate real-world dynamics, sensor noise, latencies, and environmental uncertainties. Methods such as system identification \citep{sontakke2023residual} and domain randomization \citep{tobin2017domain, peng2018sim} have been widely adopted to address this problem. \cite{learning_in_sim} proposed an \gls{rl} framework for miniature race cars that leveraged model randomization, policy regularization, and online fine-tuning of the learned policy to diminish the \gls{s2r} gap, achieving near \gls{mpc}-level performance. Despite these advances, \gls{s2r} methods often require extensive pre-training in simulation and substantial fine-tuning on physical systems, which can be resource-intensive and time-consuming and do not reach the performance of classical \gls{sota} controllers. 
Our proposed architecture bypasses these problems entirely by learning directly on-board.
% Our proposed on-car learning architecture bypasses the \gls{s2r} gap by learning directly in the target environment, eliminating the need for pre-training, system identification and manual fine-tuning.
% It is efficient enough to run locally on a single CPU, while surpassing classical controllers in terms of minimum lap times.

% \textbf{Residual Control Architectures} have been proposed to integrate classical control methods with \gls{rl}, leveraging the strengths of both model-based and model-free approaches. In this framework, an agent learns a residual policy that adjusts or refines the actions of a pre-existing base controller, allowing for improved performance and adaptability \citep{johannink2019residual}.
\textbf{Residual \gls{rl}} \citep{johannink2019residual} proposes a structure in which an agent learns a residual policy that adjusts or refines the actions of a pre-existing base controller, allowing for improved performance and adaptability. 
\cite{trumpp2023residual} introduced a residual policy learning approach specifically for \gls{ar} in simulation.
% They combined a \gls{pp} baseline controller with a residual  policy to handle unmodeled dynamics and improve performance in aggressive driving scenarios within simulation.
Their method involved training the residual policy to adjust a base \gls{pp} controller's actions and demonstrated to be effective in high-speed racing tasks by reducing lap times and enhancing stability when evaluated on simulated F1TENTH racing tracks.
% The residual agent was able to compensate for modeling inaccuracies and adapt to complex vehicle dynamics, demonstrating the effectiveness of the method in simulated high-speed racing tasks.
\cite{ghignone2025rlpp} deployed a similar approach on the physical F1TENTH platform. Such an agent, trained in simulation, was then able to improve the lap time of the base controller with the help of an \gls{rl} policy. However, the achieved lap times were still significantly slower than \gls{sota} classical algorithms.

In this work, the residual \gls{rl} approach is adopted due to its faster convergence to fast lap times, essential for achieving on-board learning with limited time.

\textbf{On-Robot learning} eliminates the \gls{s2r} gap entirely by training directly on physical robots but needs particularly efficient structures to achieve high performance \citep{daydreamer}, as the massively parallel architectures that are available in simulation \citep{rudin2021learning} are often not replicable.
Some works have been developed directly on-board, such as in \cite{train_in_austria}, where an agent could successfully navigate an oval track with three hours of training, relying solely on \gls{lidar} scans for its observations. The final policy achieved speeds of approximately 2\,m/s.
Similarly, \cite{bypassing_sim_to_real} trained an agent under the supervision of a safety system based on a viability kernel designed to ensure crash-free operation.
% This safety system employed a viability kernel to evaluate the safety of the agent's chosen actions and replaced unsafe actions with a fallback policy derived from a \gls{pp} controller. 
With this setup, the agent outperformed a simulation-trained baseline in terms of lap times and reliability, achieving efficient training within 10\,min. However, the approach was constrained by a speed limit of 2\,m/s and required the vehicle to pause every 20 steps to update the policy, highlighting areas for improvement in achieving continuous, high-speed training.
FastRLAP \citep{fastrlap} introduced an online \gls{rl} framework for \gls{ar}, achieving stable driving behavior within 20\,min of real-world training with an image-to-action approach. However, the approach required an external workstation for policy updates, constraining its scalability and applicability in resource-constrained environments, and was limited to speeds up to 4.5\,m/s.
%In contrast to our work, \cite{fastrlap} use an image-to-action approach while we utilize the state information of an onboard localization system and achieve speeds up to 7\,m/s.
\cite{pan2020imitation} used imitation learning on a scaled car to match an \gls{mpc} expert used to generate data. However, this approach did not surpass the \gls{sota} performance and depended on image-to-action methods and an onboard \gls{gpu}.

% Instead of imitating a baseline controller, our approach fine-tunes it with a learned residual policy. The framework runs directly on the car, utilizing onboard state estimation and prior information about the track's layout. It updates its policy in real-time, achieves speeds of up to 7\,m/s and faster lap times than classical controllers, while avoiding the usage of a GPU or a remote workstation.
In contrast to the previous work, our contribution consistently surpasses the \gls{sota} \gls{ar} agents, both with and without the residual structure, and displays agile racing, without any speed limitation and reaching over \qty{6.5}{\metre\per\second}. 
Similarly to other algorithms, our algorithm needs around 20\,min of training, but it only trains locally (on the scaled autonomous platform) and with no access to a \gls{gpu}.
Finally, our proposed solution is validated on a real platform across multiple track configurations and tire conditions, highlighting the empirical robustness of the method.

%%%%%%%%%%%%%%%%%%%%%%%%%%%%%%%%%%%%%%%%%%%%%%%%%%%%%%%%%%%%%%%%
%% Section: Methodology
%%%%%%%%%%%%%%%%%%%%%%%%%%%%%%%%%%%%%%%%%%%%%%%%%%%%%%%%%%%%%%%%
\section{Methodology}
\label{sec:method}

The residual control architecture utilized throughout this work is illustrated in \Cref{fig:architecture}. Detailed descriptions of each block are given in the subsequent sections.

This study is conducted and validated on the F1TENTH platform \citep{f1tenth}, a scaled racing platform with form factor 1:10. Localization and state estimation on a mapped track layout is based on a \gls{slam} system using a 2D \gls{lidar} sensor and \gls{imu}. The hardware setup, software architecture (implemented using the \gls{ros}), and the localization pipeline closely replicate the approach outlined in \cite{JFRpaper}. All computations are conducted on the onboard Intel i7-10710U processor.

\subsection{State Machine}
\label{sec:state_machine}
To facilitate efficient and autonomous training without human intervention, a state machine is implemented to recover the vehicle from terminal states. Two terminal states are defined: the first occurs when a track-boundary violation is detected, which happens if the center of the rear axle moves outside the defined track boundaries. The second terminal state is triggered by the intervention of the relative-heading safety filter, as detailed in the subsequent section.

Upon reaching a terminal state, the \gls{rl} agent is deactivated and the baseline controller steers the car back to the reference trajectory. Once the vehicle orientation realigns with the reference trajectory, a new learning episode is initiated, and the residual controller resumes operation. This process ensures seamless episode transitions and continuous data collection, enabling stable and efficient training.

\subsection{Curriculum Safety Filter}
\label{sec:safety}
A relative-heading safety filter is implemented to prevent the vehicle from leaving the track with a large relative heading, $\Delta \psi$, with respect to the reference trajectory. The safety filter signals a terminal state if $|\Delta \psi| > \psi_{\text{filter}}$, where $\Delta \psi = \psi - \psi_{\text{ref}}$, $\psi$ is the vehicle's current heading, and $\psi_{\text{ref}}$ is the heading of the reference trajectory in the global frame. The safety filter is implemented with an adaptive threshold, $\psi_{\text{filter}}$. If the agent completes a lap successfully, $\psi_{\text{filter}}$ is increased by $\epsilon$; conversely, it is decreased by the same amount if a track-boundary violation is detected.

This mechanism serves two purposes: first, it prevents the vehicle from traversing the run-off area with a heading towards the physical barriers, giving the state machine more time to intervene. Second, it facilitates the agent's learning process by acting as a curriculum mechanism \citep{bengio2009curriculum}, forcing it to follow the reference trajectory more closely in the beginning, before relaxing the constraint.

\subsection{Action Space}
\label{sec:action}
The control command is defined as $u = [\delta, v]^T$, where $\delta$ is the steering angle, and $v$ the velocity. The command is the sum of the base controller's output, $u_{\text{base}} = [\delta_{\text{base}}, v_{\text{base}}]^T$, and the \gls{rl} controller's output, $u_{\text{RL}} = [\delta_{\text{RL}}, v_{\text{RL}}]^T$.

The base controller operates within the ranges $\delta_{\text{base}} \in [-0.42, 0.42]\,\mathrm{rad}$ and $v_{\text{base}} \in [0, 10]\,\mathrm{m/s}$. The learned policy refines $u_{\text{base}}$ with residual adjustments, constrained to $\delta_{\text{RL}} \in [-0.15, 0.15]\,\mathrm{rad}$ and $v_{\text{RL}} \in [-0.5, 2]\,\mathrm{m/s}$. This formulation allows the residual controller to make fine adjustments to the base controller's output, enabling greater adaptability, precision, and ultimately faster lap times.

\subsection{Observation Space}
\label{sec:observation}
All observations are based on the onboard state estimation. The observation space is constructed similarly to \cite{tc-driver}, incorporating vehicle state, dynamics, and track information. The vehicle dynamics and state are represented by $o_{\text{car}} = [v_x, v_y, \dot{\psi}, \Delta d, \Delta \psi, \delta_{\text{base}}, v_{\text{base}}, \delta_{\text{RL,\,prev}}, v_{\text{RL,\,prev}}]^T$, where $v_x$ and $v_y$ are the lateral and longitudinal velocities in the car frame, $\dot{\psi}$ is the yaw rate, $\Delta d$ is the car's lateral deviation from the reference line, and $\Delta \psi$ is its relative heading to the reference line. $\delta_{\text{base}}$ and $v_{\text{base}}$ are the outputs of the base controller, and $\delta_{\text{RL,\,prev}}$ and $v_{\text{RL,\,prev}}$ are the outputs of the \gls{rl} controller of the previous time step.

To capture track information, the observation space includes $o_{\text{track}}$, which summarizes the upcoming reference trajectory and track boundaries. A total of $J$ equally spaced points along the reference trajectory are sampled within a horizon of $l$ meters. At each sampled point, the positions of the reference line, left boundary, and right boundary are stored in the car frame. Formally, this is expressed as
$o_{\text{track}} = [ \mathbf{p}_{\text{ref}}^0, \dots, \mathbf{p}_{\text{ref}}^J,\ \mathbf{p}_{\text{left}}^0, \dots, \mathbf{p}_{\text{left}}^J,\ \mathbf{p}_{\text{right}}^0, \dots, \mathbf{p}_{\text{right}}^J ]^T$, where $\mathbf{p}_{\text{ref}} \in \mathbb{R}^2$ represents a two-dimensional point on the race line, and $\mathbf{p}_{\text{left}}$ and $\mathbf{p}_{\text{right}}$ represent points on the left and right track boundaries, respectively. The complete observation space is then defined as $o = [o_{\text{car}}, o_{\text{track}}]^T$, with~$o \in \mathbb{R}^{5 + 6J}$ and is normalized. A graphical representation of the observation space is available in the supplementary material, \Cref{fig:suppl_ipz_overlay}.

\subsection{Reward Function}
\label{sec:reward}
The reward consists of a positive term $\lambda\Delta s$, granted at every step for progress along the reference line $\Delta s$ and scaled by the constant $\lambda$ \citep{learning_in_sim, tc-driver}, as well as a negative penalty $p$, applied only upon track-boundary violations or safety filter interventions. Both terms are mutually exclusive.

\subsection{Training Architecture}
\label{sec:architecture}
% To improve data efficiency, we extend \gls{sac} \citep{haarnoja2018soft} from Stable Baselines \citep{stable-baselines} with architectural enhancements. We adopt an asynchronous pipeline that separates the agent into a behavior actor, which collects data and fills the replay buffer, and a training actor, which updates the policy and value functions \citep{async}. Periodically, the behavior policy synchronizes with the updated training policy. Additionally, we incorporate multi-step \gls{td} learning \citep{sutton1998introduction} to leverage long-horizon rewards and better handle delayed effects, improving performance in sparse reward or in our case penalty settings.
To improve data efficiency, we extend \gls{sac} \citep{haarnoja2018soft} from Stable Baselines \citep{stable-baselines} with an asynchronous pipeline that separates the agent into an acting policy actor and a training policy, akin to \emph{Async-\gls{sac}-1} from \cite{async}. Then the proposed approach incorporates multi-step \gls{td} learning \citep{sutton1998introduction}, similarly to previous \gls{ar} work \citep{superhuman_rl}, to stabilize training \citep{barth-maron2018distributional}.

Additionally, a \gls{hdra} mechanism, inspired by principles from \gls{her} \citep{HER} and reward shaping \citep{reward_shaping}, has been adopted and integrated. In scenarios where the agent reaches a terminal state with a negative reward, the $N$ most recent transitions in the replay buffer are retroactively modified to amplify the penalty and create a stronger learning signal to discourage failure and the actions leading up to it.
% Specifically, rewards for the last $N$ steps preceding a failure are adjusted downward, creating a stronger learning signal to discourage failure and the actions leading up to it.
% This modification biases the learning process toward more robust behaviors in challenging conditions and further enhances the data efficiency of the \gls{sac} algorithm.
The stored rewards are then adjusted as follows: given an index $i_{term} \in \mathbb{N}$ corresponding to a terminal state stored in the replay buffer, the reward associated with the action that led to this state is $r_{i_{term}-1} = p$. The $N$ previous rewards ($r^{old}$) are then decreased with a linearly decreasing term as follows:
\begin{equation}
    r_{i_{term}-1-j} = r_{i_{term}-1-j}^{old} - \frac{N-j}{N}p,\quad\forall j \in [0,\,1,\,\dots,\,N-1].
\end{equation}
The corresponding pseudocode is attached in \Cref{appendix:dra}. 
% \todo{needs to enter the paper main body, possibly with equations}.

During training, the control frequency of the residual \gls{rl} controller is set to \qty{10}{\hertz}, with the training policy updated at \qty{32}{\hertz} and the behavior policy synced with the latter at \qty{1}{\hertz}. During deployment, the control frequency is increased to \qty{15}{\hertz}, while the baseline controllers operate at a fixed frequency of \qty{40}{\hertz} throughout both training and deployment phases. The output of the \gls{rl} controller is continuously added to the output of the base controller.

%%%%%%%%%%%%%%%%%%%%%%%%%%%%%%%%%%%%%%%%%%%%%%%%%%%%%%%%%%%%%%%%
%% Section: Evaluation
%%%%%%%%%%%%%%%%%%%%%%%%%%%%%%%%%%%%%%%%%%%%%%%%%%%%%%%%%%%%%%%%
\section{Experimental Setup}
\label{sec:setup}

The effectiveness and robustness of the proposed architecture are evaluated through a series of time-trial experiments. 
The primary objective is to minimize lap time while maintaining stability and reliability over multiple laps.
Each scenario is allocated a single \qty{5000}{\milli\ampere\hour} battery for training, which lasts between \num{20} and \qty{33}{\min} depending on the track-tire configuration and speed profile.
After training, the agent is deployed until it completes \num{20} laps without track-boundary violations.
Metrics such as lap times and number of boundary violations are recorded to evaluate speed, repeatability, and adaptability.

\subsection{Evaluation}
\label{sec:evaluation}

One of the most important goals of this evaluation is to demonstrate the generalization of the proposed approach. This is done by integrating the residual \gls{rl} controller with three distinct baseline controllers (\gls{pp}, \gls{map}, and \gls{ftg}, as detailed in \Cref{sec:baselines}) and by testing it in an \gls{e2e} configuration. Each configuration was evaluated on the \qty{41}{\metre} long \texttt{C-track}, with performance measured by the minimum lap time over \num{20} consecutive deployment laps.

Following these initial experiments, the best-performing combination was selected for further evaluation. These extended tests included repeated training and deployment with varying random seeds on the \texttt{C-track} and on the \qty{34}{\metre} long \texttt{Y-track} to assess repeatability and robustness. While these primary experiments were conducted using Turbo tires with a relatively high static friction coefficient of $\mu^{Turbo}_{static} \approx 1.01$, additional experiments were performed on the \texttt{C-track} using \gls{tpu} tires, which exhibit a lower static friction coefficient of $\mu^{TPU}_{static} \approx 0.75$ and different cornering behavior. This allowed for a comprehensive analysis of the controller's performance under varying track and tire conditions. Images of the F1TENTH car, tires, and track setup can be seen in \Cref{appendix:setup}.

\subsection{Baselines}
\label{sec:baselines}

The following baseline controllers are used to evaluate the agent's learning enhancements and integration capabilities with different baseline controllers:

\textbf{\gls{pp}}: A classical geometric controller that calculates steering based on a look-ahead point on a pre-defined reference trajectory. This controller operates under a no-slip assumption, making it limited at high speeds where friction dynamics become significant \citep{pure_pursuit}.

\textbf{\acrfull{map}}: An extended pursuit method that integrates a lookup table to incorporate tire dynamics, allowing for improved handling in high-speed corners by accounting for non-linear friction effects \citep{map}.

\textbf{\acrfull{ftg}}: This reactive approach identifies gaps in the \gls{lidar} measurements to navigate through \citep{ftg}. Therefore, it is often the chosen controller when navigating unknown environments with physical boundaries. However, due to its reactive behavior and lack of knowledge regarding the upcoming track sections, it often lags behind in terms of lap time.

\textbf{\acrfull{e2e}}: To demonstrate the data efficiency of the proposed \gls{rl} architecture and isolate its performance, experiments without a baseline controller are conducted. The range for the commanded steering angle is extended to $[-0.42;0.42]$ rad, which aligns with the limits of the physical steering actuator. For the commanded speed, a curriculum learning approach has been implemented \citep{curriculum_learning}. The action space for the speed command is $[0.5;\alpha]$ with $\alpha \in [1;7]$\,m/s. $\alpha$ is increased by 0.5\,m/s as soon as the agent completes three consecutive laps without track-boundary violation. Naturally, the baseline control command is removed from the observation space.

\textbf{Remark}: While \gls{mpc} provides precise trajectory tracking and constraint adherence, it is not combined with the residual controller in this study as its constraint-based nature inherently conflicts with \gls{rl}’s adaptive learning, especially when the agent attempts to operate near the limits of the \gls{mpc}’s predefined constraints. Nonetheless, we include \gls{mpc} performance in our baseline evaluation, as it is recognized as one of the most performant controllers in the literature \citep{vazquez2020mpccurv} and provides a valuable reference point for contextualizing our results.
More detail on the optimization problem formulation, also derived from the work of \cite{vazquez2020mpccurv}, is available in \Cref{appendix:mpc_formulation}.

%%%%%%%%%%%%%%%%%%%%%%%%%%%%%%%%%%%%%%%%%%%%%%%%%%%%%%%%%%%%%%%%
%% Section: Results
%%%%%%%%%%%%%%%%%%%%%%%%%%%%%%%%%%%%%%%%%%%%%%%%%%%%%%%%%%%%%%%%
\section{Results}
\label{sec:results}

\begin{figure}[ht]
    \begin{center}
        \includegraphics[width=\textwidth]{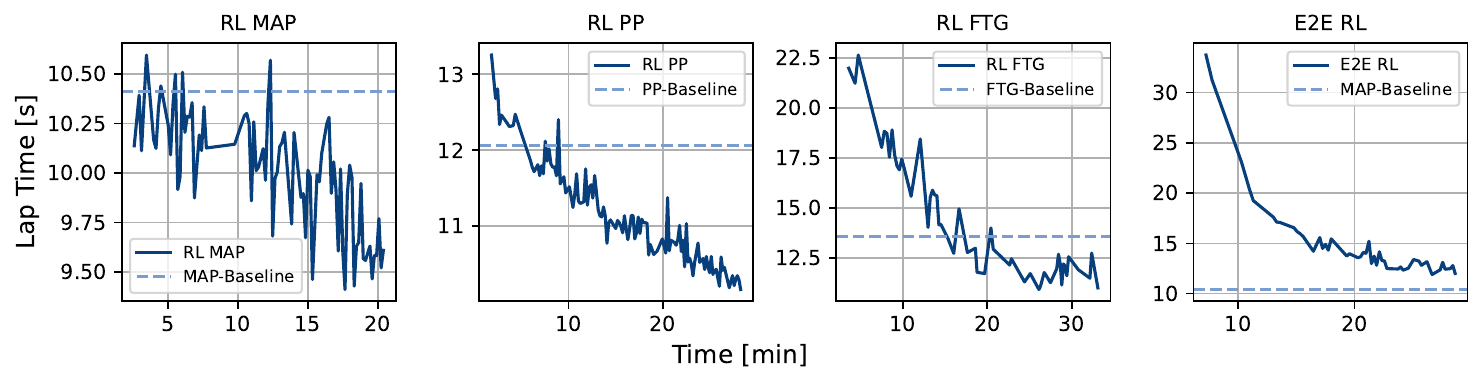}
    \end{center}
    \caption{Lap Time vs. Wall Time during training for different residual controllers, with respective reference for best lap time of the baseline controller and the \gls{e2e} controller with \gls{map} baseline. Only lap times without boundary violations are recorded.}
    \label{fig:ctrl_comp}
\end{figure}

\Cref{fig:ctrl_comp} shows the lap time improvement during training for the different residual \gls{rl} controllers and the \gls{e2e} \gls{rl} controller on the 
\texttt{C-track}, highlighting the agent's ability to improve upon the performance of all baseline controllers.
While \gls{rl} \gls{map} and \gls{rl} \gls{pp} outperformed their respective base controller within \qty{10}{\minute}, \gls{rl} \gls{ftg} needed about \qty{20}{\minute} to achieve the same. The \gls{e2e} agent also shows significant lap time improvements within \qty{20}{\minute}, but does not manage to reach the \gls{sota} baseline set by the \gls{map} controller.

During deployment, the \gls{rl} \gls{map} controller achieved the fastest lap time of 9.26\,s, significantly surpassing the \gls{map} baseline’s best of 10.41\,s. 
Across all baseline-residual \gls{rl} combinations, lap time reductions ranged from 11.0\,\% to 20.5\,\% for best lap times and 9.0\,\% to 20.3\,\% for mean lap times. An extended table listing all minimum, average and the \gls{sd} of the lap times for the different combinations can be seen in \Cref{sec:app_base_controller}. \gls{rl} \gls{map}’s superior results in terms of minimum and average lap time made it the primary choice for subsequent experiments.

\textbf{E2E extended training}: Since training the \gls{e2e} controller for only one battery life was not sufficient to obtain \gls{sota} performance, the experiment was extended to three battery lives, cumulating in \qty{82}{\minute} of on-track learning. Afterward, the agent's average lap time was \qty{5.0}{\percent} lower than the \gls{sota} set by the \gls{map} controller, while the minimum lap time was \qty{7.3}{\percent} lower (Extended \gls{e2e} \gls{rl}: $t_{min}=$\qty{9.65}{\second}, $t_\mu=$\qty{9.97}{\second}. \gls{map}: $t_{min}=$\qty{10.41}{\second}, $t_\mu=$\qty{10.50}{\second}). During the deployment, only 4 track-bound violations and speeds up to \qty{7}{\metre \per \second} were registered. Extended data and visualizations of the deployment runs can be seen in \Cref{sec:app_base_controller}.

\subsection{\gls{rl} \gls{map} Extended Evaluation}
\label{sec:rl_map}
The \gls{rl} \gls{map} controller was further re-trained from scratch in two different conditions and across three different random seeds to evaluate the method across environments. 
The first environment consisted of using the more slippery \gls{tpu} tires on the same track (\texttt{C-track}), while the second experiment evaluated the algorithm on the \texttt{Y-track}.
Tires pictures and track layouts are available in \Cref{appendix:setup} and \Cref{sec:app_scratch}.
The \gls{rl} \gls{map} controller consistently outperformed its baseline across all scenarios tested, achieving improvements ranging from 5.4\% to 11.5\% for minimum lap times and 5.9\% to 9.0\% for mean lap times. Even during the worst deployment run in terms of lap time reduction, the proposed architecture managed to decrease the minimum lap time by 0.57\,s and the average lap time by 0.51\,s. This highlights the robustness of the approach across varying track layouts, friction levels, and weight initialization of the neural network. \Cref{tab:deploy_scratch} summarizes the three deployment runs per tire and track combination. A detailed break-down of the individual deployments is presented in the \Cref{sec:app_scratch}.

\begin{table}[h!]
\centering
\small % Reduce font size slightly
\setlength{\tabcolsep}{6pt} % Adjust space between columns
\renewcommand{\arraystretch}{1.2} % Adjust row spacing
\resizebox{\textwidth}{!}{ % Resize the table to fit within the text width
\begin{tabular}{lcccccccccccc}
\toprule
\multicolumn{1}{c}{} & \multicolumn{4}{c}{C-Track, Turbo} & \multicolumn{4}{c}{C-Track, TPU} & \multicolumn{4}{c}{Y-Track, Turbo} \\
\cmidrule(lr){2-5} \cmidrule(lr){6-9} \cmidrule(lr){10-13}
Ctrl. & $t_{\text{min}}$ ↓ & $t_{\mu}$ ↓ & $\sigma$ ↓ & $n_{\text{b}}$ ↓ & $t_{\text{min}}$ ↓ & $t_{\mu}$ ↓ & $\sigma$ ↓ & $n_{\text{b}}$ ↓ & $t_{\text{min}}$ ↓ & $t_{\mu}$ ↓ & $\sigma$ ↓ & $n_{\text{b}}$ ↓ \\
\midrule
MPC       & 11.86 & 11.95 & 0.083 & \textbf{0} & 13.14 & 13.25 & 0.179 & \textbf{0} & 9.97 & 10.04 & 0.071 & 1 \\
\gls{map}       & 10.41 & 10.50 & \textbf{0.060} & \textbf{0} & 12.93 & 13.45 & 0.338 & \textbf{0} & 8.55 & 8.62 & \textbf{0.045} & \textbf{0} \\
\gls{rl} \gls{map} (Ours)   & \textbf{9.26} & \textbf{9.56} & 0.181 & 2 & \textbf{12.23} & \textbf{12.50} & \textbf{0.163} & 1 & \textbf{7.57} & \textbf{7.97} & 0.205 & 6 \\
\bottomrule
\end{tabular}
}
\caption{Performance of \gls{rl} \gls{map} with different tire-track combinations averaged over three randomly initialized seeds compared to \gls{map} and \gls{mpc} for reference. $t_{\text{min}}$ [s] represents the minimum lap time, $t_{\mu}$ [s] the average lap time, $\sigma$ [s] the standard deviation of the lap times and $n_{\text{bound}}$ the number of boundary violations until 20 violation-free laps are reached.}
\label{tab:deploy_scratch}
\end{table}

In terms of lap time repeatability, \gls{rl} \gls{map} exhibited a more than three times larger lap time standard deviation compared to its baseline controller on the high-friction Turbo tires. However, in the low-friction scenario, \gls{rl} \gls{map} demonstrated greater consistency with an \gls{sd} of 0.16\,s, producing more stable lap times than its baseline controller with a SD of 0.34\,s. While the \gls{map} baseline was capable of completing 20 consecutive laps without any track-boundary violations across all scenarios, the residual agent experienced up to two violations on the \texttt{C-track} with Turbo tires, up to one violation on the \texttt{C-track} with TPU tires, and up to five on the \texttt{Y-track} track before successfully completing 20 violation-free laps. The majority of these violations occurred due to slight corner clipping.

\begin{figure}[H]
    \centering
    % First figure
    \begin{subfigure}{0.32\textwidth}
        \centering
        \includegraphics[width=\textwidth]{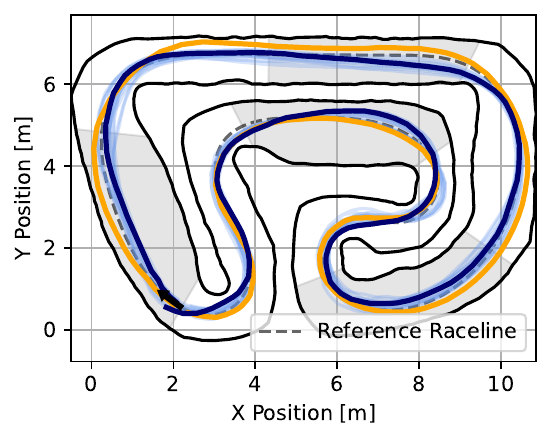}
    \end{subfigure}
    \hfill
    % Second figure
    \begin{subfigure}{0.32\textwidth}
        \centering
        \includegraphics[width=\textwidth]{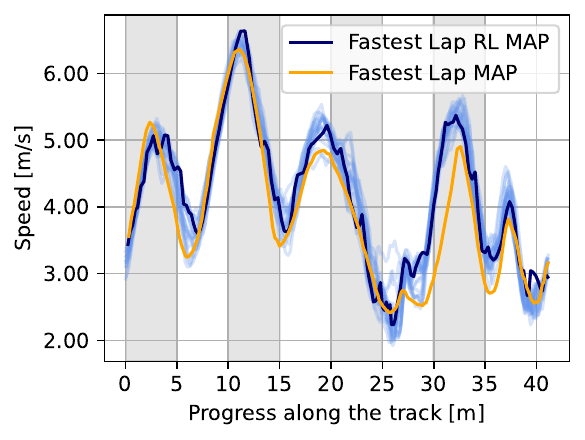}
    \end{subfigure}
    \hfill
    % Third figure
    \begin{subfigure}{0.32\textwidth}
        \centering
        \includegraphics[width=\textwidth]{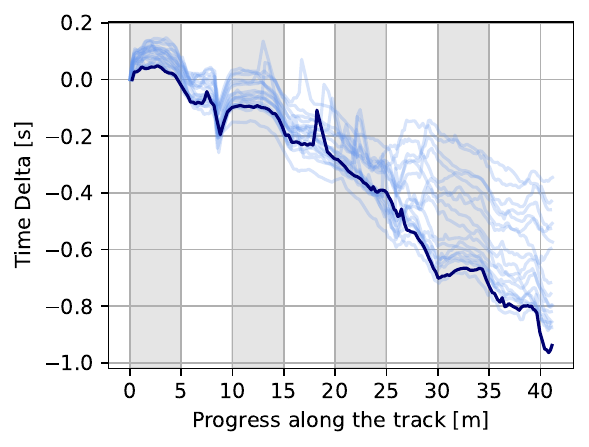}
    \end{subfigure}

    \caption{Comparison of the driven trajectories, speed profiles, and the time delta for \gls{rl}~\gls{map}~(blue) vs. \gls{map} (orange) on the \texttt{C-track} with Turbo tires.
    Speed profiles and time deltas are represented along the positional advancement on the track.
    The zero progress position corresponds to the black arrow in the bottom left corner, and the track is driven in the sense of the arrow.
    Dark blue represents the fastest lap of \gls{rl}~\gls{map}.}
    \label{fig:rbring_turbo_scratch}
\end{figure}

Qualitative observations indicate that the \gls{rl} \gls{map} controller adjusts steering and speed effectively across the track. For example, on the \texttt{C-track} with Turbo tires (\Cref{fig:rbring_turbo_scratch}) the residual controller negotiates all the corners but one with higher minimum speed, without sacrificing straight speed, as it also reaches higher top velocities at the end of the acceleration sections. Similar behavior can be observed in the other configurations in the supplementary material, \Cref{sec:app_scratch}.

%% GENERALIZATION
\subsection{\gls{rl} \gls{map} Generalization Evaluation}
Generalization to a different track layout was evaluated by deploying an \gls{rl} \gls{map} agent, trained on the \texttt{C-track} with Turbo tires, on the \texttt{Y-track} with same tires using zero-shot (no additional training) and few-shot transfer (additional training for one battery life on the new track).

% Seed 1 and 2 outperformed the baseline in terms of best and mean lap times. However, Seed 2 displayed instability, recording 14 track-boundary violations all in the same corner due to corner clipping. The deployment of Seed 3 had to be terminated after 20 track-boundary violations, primarily caused by repeated corner clipping. Visualizations of all deployment runs can be seen in the supplementary material in  \Cref{fig:rbring_tpu_zero_deploy_1_part1}.
% In zero-shot transfer, the results were mixed. Seed 1 and 2 showed promise, with, respectively, $t_{\mu}=...,\,\sigma=...$ and $t_{\mu}=...,\,\sigma=...$.
In zero-shot transfer, the results across random seeds were mixed and not all 20 laps could be completed, with the aggregate deployment attaining $t_\mu=8.68\,s,\,\sigma=0.34\,s$.
While Seed 1 managed to outperform the baseline in terms of best and mean lap times after only one collision, Seed 2 and Seed 3 experienced significant instability, with multiple track-boundary violations that necessitated terminating the deployment.

\begin{figure}[ht]
    \begin{center}
        \includegraphics[width=\textwidth]{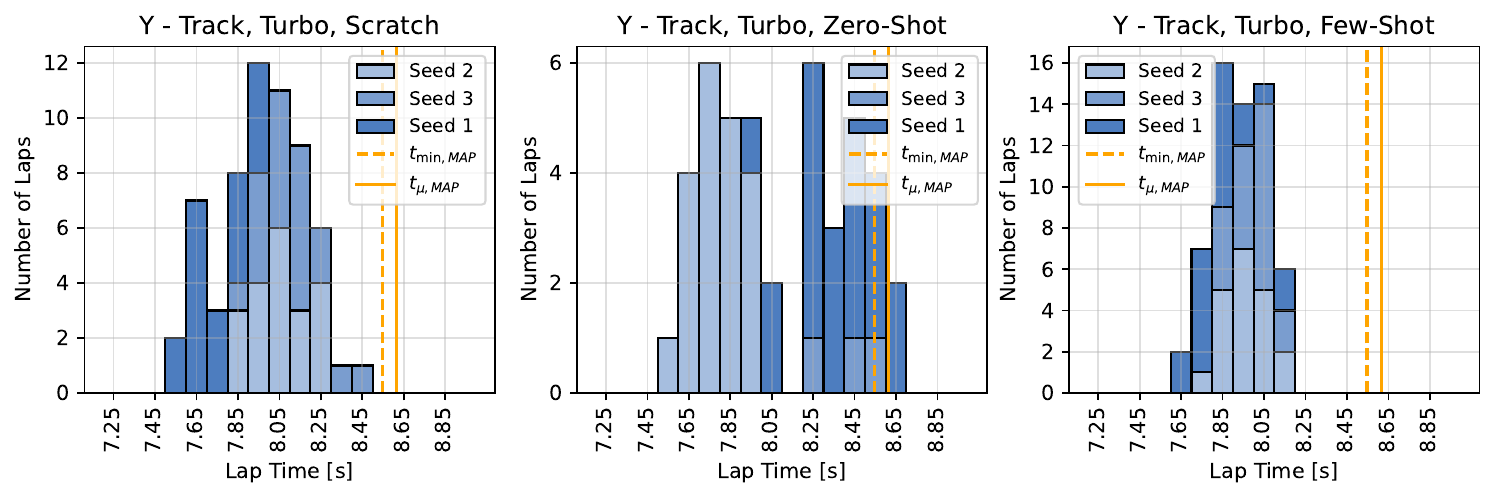}
    \end{center}
    \caption{Stacked histogram of lap times achieved during zero-shot and few-shot transfer from the \texttt{C-track} to \texttt{Y-track} during 20 boundary-violation-free laps. Different shades represent the three different weight initialization and resulting policies learned on the \texttt{C-track}.}
    \label{fig:hist_zero_few_main}
\end{figure}

Due to the very limited zero-shot transfer capabilities of the method, we proceeded to further test few-shot transfer, which involved additional training for one battery life.
This led to improvements in both lap repeatability and lap times compared to zero-shot transfer. When compared to policies trained from scratch on the \texttt{Y-track}, the few-shot transfer policy demonstrated fewer track-boundary violations (up to two per seed, four in total), an on average \qty{4.7}{\percent} lower mean lap time and half the standard deviation in lap times across all random seeds.
The aggregate laps demonstrated $t_\mu=$\qty{7.93}{\second}, $\sigma=$\qty{0.13}{\second}.
\Cref{fig:hist_zero_few_main} presents a distribution of all the lap times recorded for the evaluation, highlighting the effectiveness of few-shot transfer in enhancing lap repeatability.
Extended results of the generalization tests can be found in \Cref{sec:app_generalization}.
%% END GENERALIZATION EXTRA TEXT

\subsection{Ablation Study}
\label{sec:ablation}
\begin{wrapfigure}{r}{0.5\textwidth} % 'r' for right, 'l' for left
  \centering
  \includegraphics[width=0.5\textwidth]{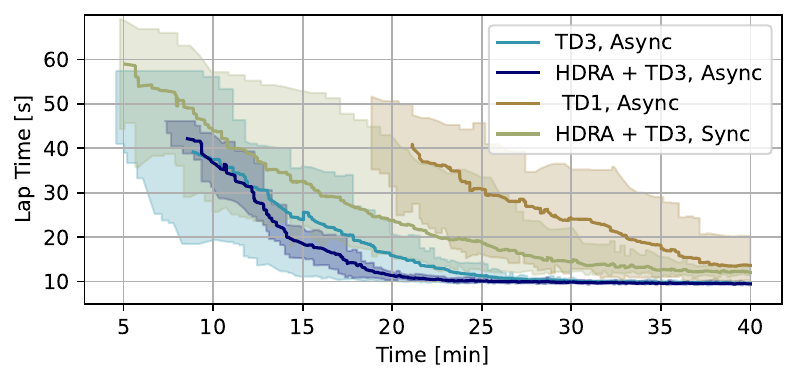}
  \caption{Lap times during training averaged over 5 runs for ablation regarding multi-step \gls{td} over one (TD1) and three (TD3) steps, a synchronous (sync) and asynchronous (async) architecture as well as with \gls{hdra} and without. Lines only start when the first collision-free lap is completed. The area between the minimum and the maximum lap time is shaded.}
  \label{fig:ablation}
\end{wrapfigure}
We evaluate the impact of key architectural and algorithmic modifications with an ablation study conducted in the real-time F1TENTH ROS simulator \citep{simulator}. This study focuses on the more challenging \gls{e2e} setting, comparing various combinations of multi-step \gls{td} learning, \gls{hdra}, and training architectures (asynchronous vs. synchronous). \Cref{fig:ablation} presents the results, averaged over five independent training runs with different random seed initializations.

A first comparison shows that in our experiments the asynchronous architecture (\gls{hdra} + TD3, Async) manages to reduce the lap times faster than the synchronous architecture (\gls{hdra} + TD3, Sync), and in the same training time achieves a faster lap time compared to its synchronous counterpart. This improvement is likely due to the ability of the asynchronous architecture to support a policy update rate three times higher.

Furthermore, the addition of multi-step \gls{td} error (TD3, Async), shows in our experiments to help speed up the learning process, achieving lap completions earlier than the ten-minute mark, while (TD1, Async) needs in this case around \qty{20}{min}.
Finally, the addition of \gls{hdra} (\gls{hdra} + TD3, Async) does not show a statistically relevant improvement against (TD3, Async) in average lap time, but keeps the lap time spread lower, and makes the training more consistent across random seeds.
% The study reveals that combining \gls{hdra} with three-step \gls{td} error in an asynchronous architecture (\gls{hdra} + TD3, Async) exhibits 
% % greater robustness to seed initialization and achieves sub-ten-second lap times on average \qty{5}{\minute} earlier than TD3 alone.
% narrower spread of lap times, and decreases the maximum lap times faster than TD3 alone.
% Additionally, the asynchronous architecture not only accelerates the overall lap time reduction but also achieves, on average, a \qty{2.5}{\second} faster lap time compared to its synchronous counterpart. This improvement is likely due to the ability of the asynchronous architecture to support a policy update rate three times higher. These findings highlight the critical role of multi-step \gls{td} learning and asynchronous training in improving the time until the policy converges, while \gls{hdra} plays a smaller but meaningful role in stabilizing the training process.
Detailed metrics on the lap times achieved during the last three minutes of training are available in \Cref{appendix:abl}.

%%%%%%%%%%%%%%%%%%%%%%%%%%%%%%%%%%%%%%%%%%%%%%%%%%%%%%%%%%%%%%%%
%% Section: Conclusion
%%%%%%%%%%%%%%%%%%%%%%%%%%%%%%%%%%%%%%%%%%%%%%%%%%%%%%%%%%%%%%%%
\section{Conclusion}
\label{sec:conclusion}

This work presented a real-world on-board learning framework for \gls{ar}, applied to a residual and \gls{e2e} control framework. The residual controller consistently improves the performance of its baseline across all tested scenarios, achieving substantial reductions in minimum and average lap times. Among the tested configurations, the \gls{rl} \gls{map} controller delivers the fastest lap times, improving the minimum lap time by up to \qty{11.5}{\percent} and the average lap time by up to \qty{10.0}{\percent} compared to the classical \gls{map} controller. Beyond improved performance, it demonstrates strong repeatability and stability across different tracks and tire configurations, as well as weight initializations. Moreover, the proposed asynchronous framework is highly efficient, reaching said performance within a single battery life of 20 to \qty{33}{\minute} (10\,k to 19\,k environmental steps) for training.

The \gls{e2e} \gls{rl} controller further illustrates the versatility of the architecture. After \qty{29}{\minute} of training, it navigated the track consistently at speeds up to \qty{4}{\metre\per\second} and with \qty{82}{\minute} of training, surpassed the \gls{map} baseline by over half a second, maintaining stable performance at speeds up to \qty{7}{\metre\per\second}. The ablation study shows that especially the asynchronous structure and multi-step \gls{td} learning attributes to the fast convergence to the minimal lap time.

Furthermore, while the few-shot generalization experiments show that a single policy can be tuned to drive consistently on a different track, the zero-shot experiments show that future work could focus on methods targeting generalization across different real-world conditions. 
For example, recent work on data-driven simulation \citep{xiao2024anycaranywherelearninguniversal, rothfuss2024bridging} could be a starting point for creating a generalist policy that would then benefit from the framework proposed here for on-board fine-tuning.

Overall, these findings highlight the potential of a residual \gls{rl} framework to enhance classical controllers, while the same framework enables \gls{e2e} \gls{rl} agents to achieve competitive real-world performance. Future research could focus on reducing track-boundary violations during training via a model-based safety filter \citep{evans2023, zeilinger} or extending the residual control architecture to multi-agent racing, which would introduce new challenges, such as dynamic collision avoidance and strategic adaptation. Finally, the proposed architecture could be adapted for full-scale vehicles.

\subsubsection*{Broader Impact Statement}
\label{sec:broaderImpact}
The versatility of the framework also makes it a candidate for other domains, including drone racing, robotic locomotion, and manipulation tasks. In future work, we are considering exploring how these applications could validate this work's effectiveness in different high-speed or precision-critical scenarios, expanding its relevance to a broader range of autonomous systems.

%%%%%%%%%%%%%%%%%%%%%%%%%%%%%%%%%%%%%%%%%%%%%%%%%%%%%%%%%%%%%%%%
%% NOTE: THIS MARKS THE END OF THE "MAIN TEXT"
%%%%%%%%%%%%%%%%%%%%%%%%%%%%%%%%%%%%%%%%%%%%%%%%%%%%%%%%%%%%%%%%

%%%%%%%%%%%%%%%%%%%%%%%%%%%%%%%%%%%%%%%%%%%%%%%%%%%%%%%%%%%%%%%%
%% Bibliography
%%%%%%%%%%%%%%%%%%%%%%%%%%%%%%%%%%%%%%%%%%%%%%%%%%%%%%%%%%%%%%%%
\bibliography{main}
\bibliographystyle{rlj}

%%%%%%%%%%%%%%%%%%%%%%%%%%%%%%%%%%%%%%%%%%%%%%%%%%%%%%%%%%%%%%%%
% AUTHOR: If your paper has no supplementary materials, you may 
%         comment out the line below, which creates the title for
%         the supplementary materials.
%%%%%%%%%%%%%%%%%%%%%%%%%%%%%%%%%%%%%%%%%%%%%%%%%%%%%%%%%%%%%%%%
\beginSupplementaryMaterials

%%%%%%%%%%%%%%%%%%%%%%%%%%%%%%%%%%%%%%%%%%%%%%%%%%%%%%%%%%%%%%%%
%% Appendices
%%%%%%%%%%%%%%%%%%%%%%%%%%%%%%%%%%%%%%%%%%%%%%%%%%%%%%%%%%%%%%%%
\appendix
\label{appendix}

\section{Extended Related Works}
\label{appendix:related_works}
Autonomous driving research has evolved significantly, from early experiments in the 1970s to today's sophisticated systems.  \cite{zhao2024} provide a comprehensive overview of these advancements, highlighting the field’s interdisciplinary nature. Foundational projects like the NavLab series \citep{thorpe1988} and the ARGO vehicle \citep{broggi1999} laid the groundwork for integrating technologies such as computer vision, machine learning, and advanced sensor systems into autonomous systems. These pioneering efforts were further accelerated by the DARPA Grand Challenges in 2004 and 2005, which catalyzed the adoption of machine learning and data-driven methods in autonomous vehicle development, driving the transition to computer-controlled vehicles with enhanced autonomy and intelligence \citep{thrun2006}.

Various architectures have since been explored in autonomous driving. Layered architectures, with their modular design, excel in reliability and debugging, while end-to-end systems streamline processing through neural networks, though they face challenges related to data dependency and interpretability \citep{grigorescu2020}. Significant progress has also been made in perception technologies. For example, Region-based Convolutional Neural Networks \citep{girshick2015} and single-stage methods like \gls{yolo} and the \gls{ssd} have revolutionized image-based object detection \citep{yolo, ssd}. Similarly, advancements in \gls{lidar}-based systems, including Complex-\gls{yolo} and PointPillars, have enhanced 3D perception, critical for real-time navigation and obstacle avoidance \citep{simony2018, lang2019, zhang2020}.

The evolution of \gls{slam} systems underscores the increasing emphasis on accurate and efficient localization and mapping. Diverse projects demonstrate how vision-based techniques combined with inertial navigation improve robustness across diverse environments \citep{zou2022}.

While autonomous driving technologies have matured for structured environments, recent research has shifted towards more dynamic domains like \gls{ar}. This transition necessitates not only robust perception and planning but also advanced control algorithms capable of operating at the limits of vehicle dynamics. Classical control methods, such as \gls{pp} and \gls{mpc}, have served as reliable baselines. However, \gls{rl} has emerged as a promising paradigm, offering superior adaptability and agility for racing tasks.

\textbf{Reinforcement Learning in Simulation}

In high-fidelity simulators like Gran Turismo, \gls{rl} has demonstrated exceptional performance. For instance, \citet{Fuchs2021} and \citet{superhuman_rl} outperformed human drivers using \gls{sac} and multi-step \gls{td} learning. While \citet{Fuchs2021} employed an observation space similar to ours, \citet{superhuman_rl} incorporated a front-facing camera stream. By leveraging an asymmetric \gls{sac} architecture, their approach relied on propriocentric observations during deployment, excluding information about upcoming track sections. Despite its success in simulation, this method has not yet demonstrated effective real-world transferability.

While \gls{rl} methods often outperform classical approaches also in the F1TENTH simulator, this performance fails to transfer to physical environments due to the \gls{s2r} gap, as highlighted in the works of \cite{EVANS2023100496}. Factors such as model mismatch, sensor noise, and hardware delays exacerbate this gap, leading to instability and erratic behaviors in real-world deployments. Agents that outperformed a \gls{pp} controller in simulation displayed rapid side-to-side swerving and instability during real-world tests, reaching lower speeds and ultimately lower lap times. These findings underscore the persistent challenges of \gls{s2r} transfer for \gls{rl}-based autonomous racing systems.

\textbf{Trajectory-Aided \acrlong{rl}}

Trajectory-aided \gls{rl} combines classical trajectory planning with \gls{rl} to enhance safety, stability, and performance in high-speed racing. Methods such as those proposed by \citet{tc-driver} and \citet{traj_aided_rl} enforce adherence to precomputed trajectories by penalizing deviations. In contrast, our approach primarily uses trajectory information to inform the \gls{rl} agent about upcoming track sections and define progress along the track, balancing flexibility and control in dynamic environments.

\textbf{Model-Based \acrlong{rl}}
Model-based \gls{rl} has been always well regarded for its sample efficiency, first demonstrated by \cite{deisenroth2011pilco} and more recently in the \gls{mbpo} \citep{janner2019MBPO} and Dreamer \citep{dreamer} architectures.
Both these methods were shown to work well in real-world tasks, such as for drifting maneuvers \citep{rothfuss2024bridging}, for locomotion and navigation \citep{daydreamer}, and on the F1TENTH platform, where \citet{brunnbauer2022latent} applied Dreamer to achieve faster lap times in simulation compared to model-free methods.
However, this last application exhibited bang-bang control behavior when transferred to the real world, resulting in suboptimal performance.
Additionally, model-based methods require high computational demands and external workstations, making this solution unsuitable for fully on-board learning on resource-constrained platforms like the F1TENTH one, highlighting the need for more efficient on-car learning methods.

\clearpage

\section{Experimental Setup}
\label{appendix:setup}
\begin{figure}[htbp]
    \centering
    % First figure
    \centering
    \begin{subfigure}{\textwidth}
        \includegraphics[width=\linewidth]{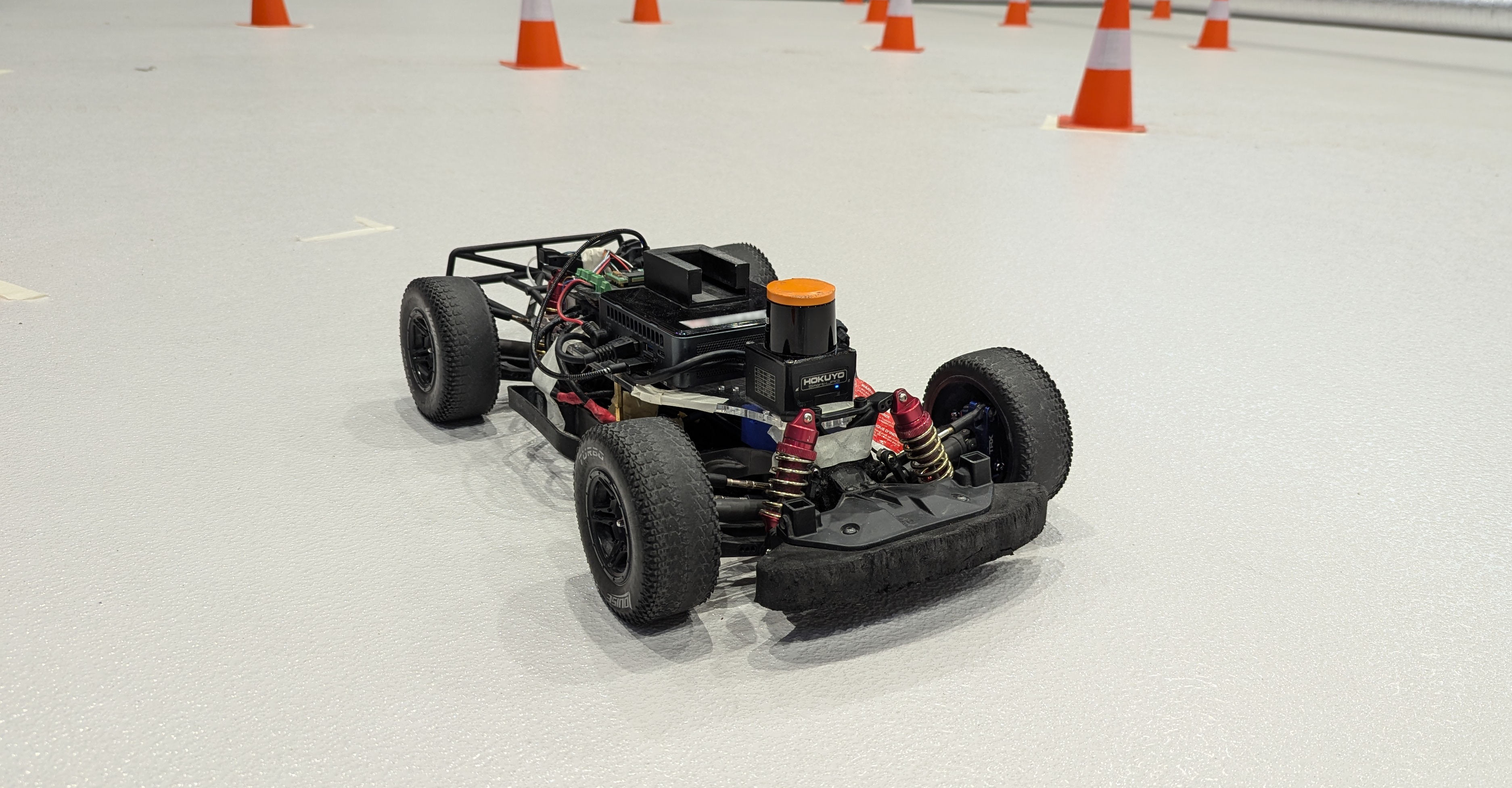}
    \end{subfigure}
    \caption{Image of the F1TENTH car used for evaluation equipped with the Intel NUC 10 computer (i7-10710U processor) and the Hokuyo UST‐10LX \gls{lidar} system.}
\end{figure}

\begin{figure}[htbp]
    \centering
    % First figure
    \centering
    \begin{subfigure}{\textwidth}
        \includegraphics[width=\textwidth]{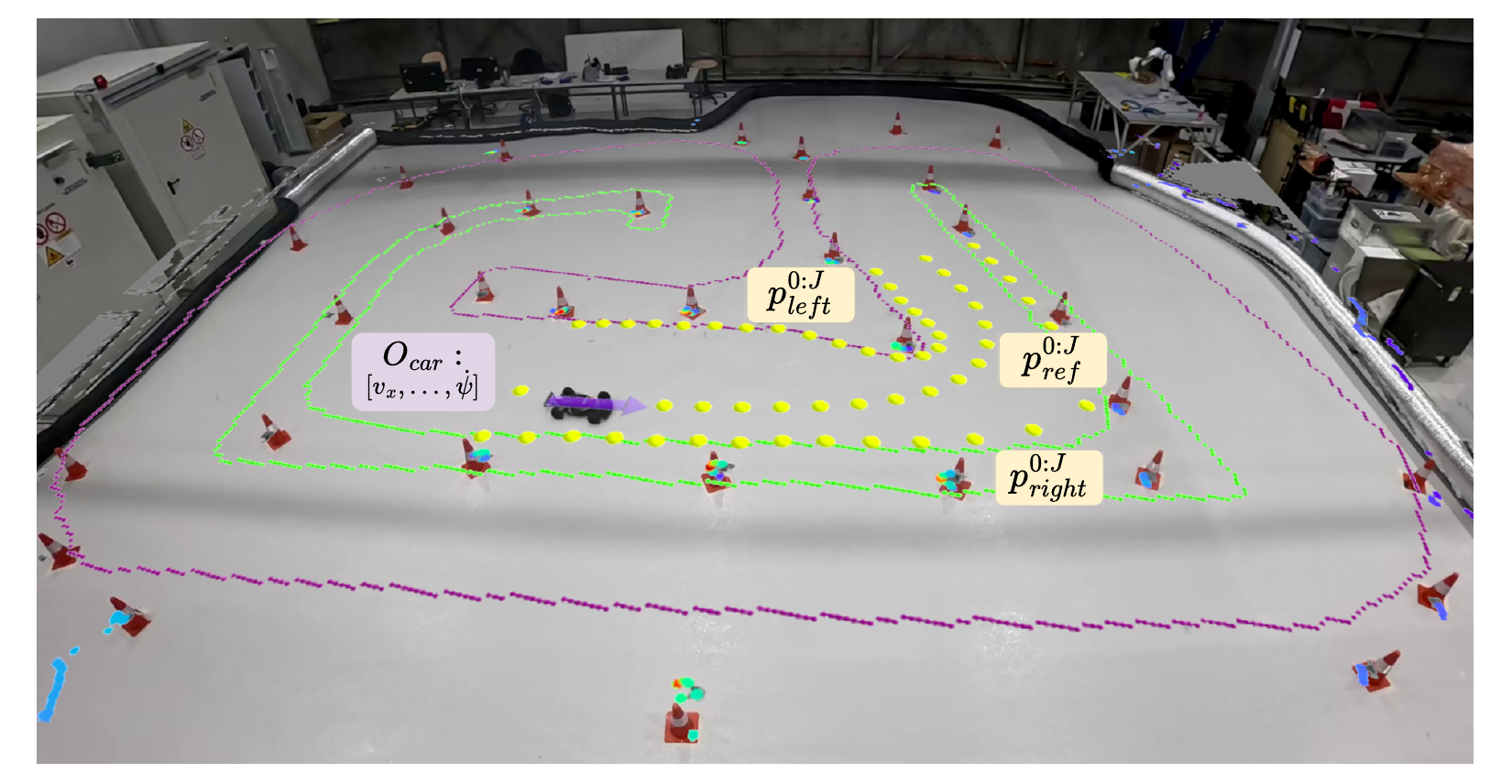}
    \end{subfigure}   
    \caption{Image of \texttt{C-track} with an overlay of the car state observation $O_{\text{car}} = [v_x, v_y, \dot{\psi}, \Delta d, \Delta \psi, \delta_{\text{base}}, v_{\text{base}}, \delta_{\text{RL,\,prev}}, v_{\text{RL,\,prev}}]^T$, denoted as shortened state vector in purple.
    The observation of the track information $O_{track} = [ \mathbf{p}^{0:J}_{\text{ref}}, \mathbf{p}^{0:J}_{\text{left}}, \mathbf{p}^{0:J}_{\text{right}}]^T$ is depicted in yellow.
    The virtual boundaries (green: inside, purple: outside), and \gls{lidar} scans are further included. Track-boundary violations during evaluation are automatically detected based on the virtual boundaries. The cones are placed as reference points for the human viewers.} 
    \label{fig:suppl_ipz_overlay}
\end{figure}

\begin{figure}[htbp]
    \centering
    % First figure
    \begin{subfigure}{0.49\textwidth}
        \centering
        \includegraphics[width=\linewidth]{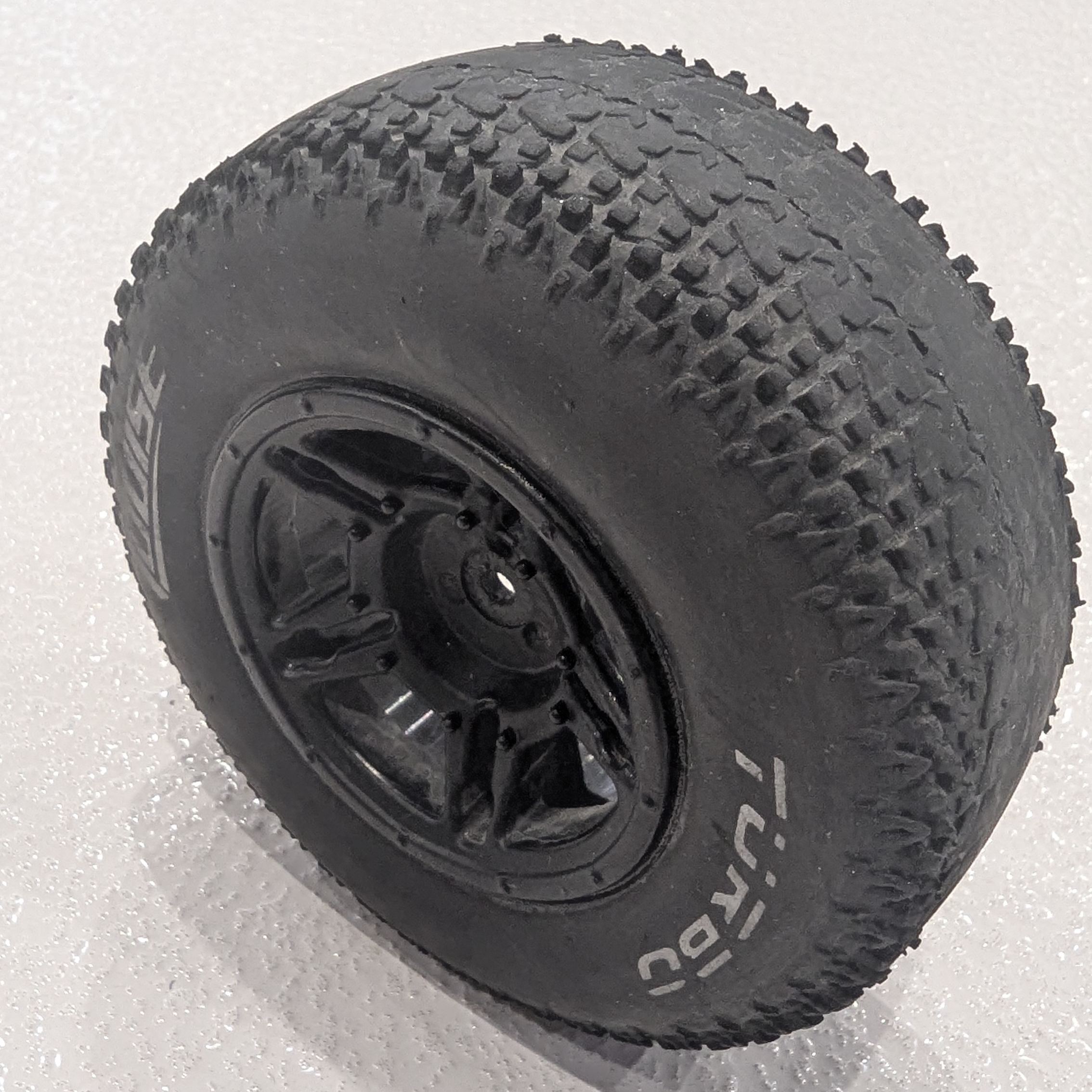}
        \caption{Turbo Tire}
    \end{subfigure}
    % Second figure
    \begin{subfigure}{0.49\textwidth}
        \centering
        \includegraphics[width=\linewidth]{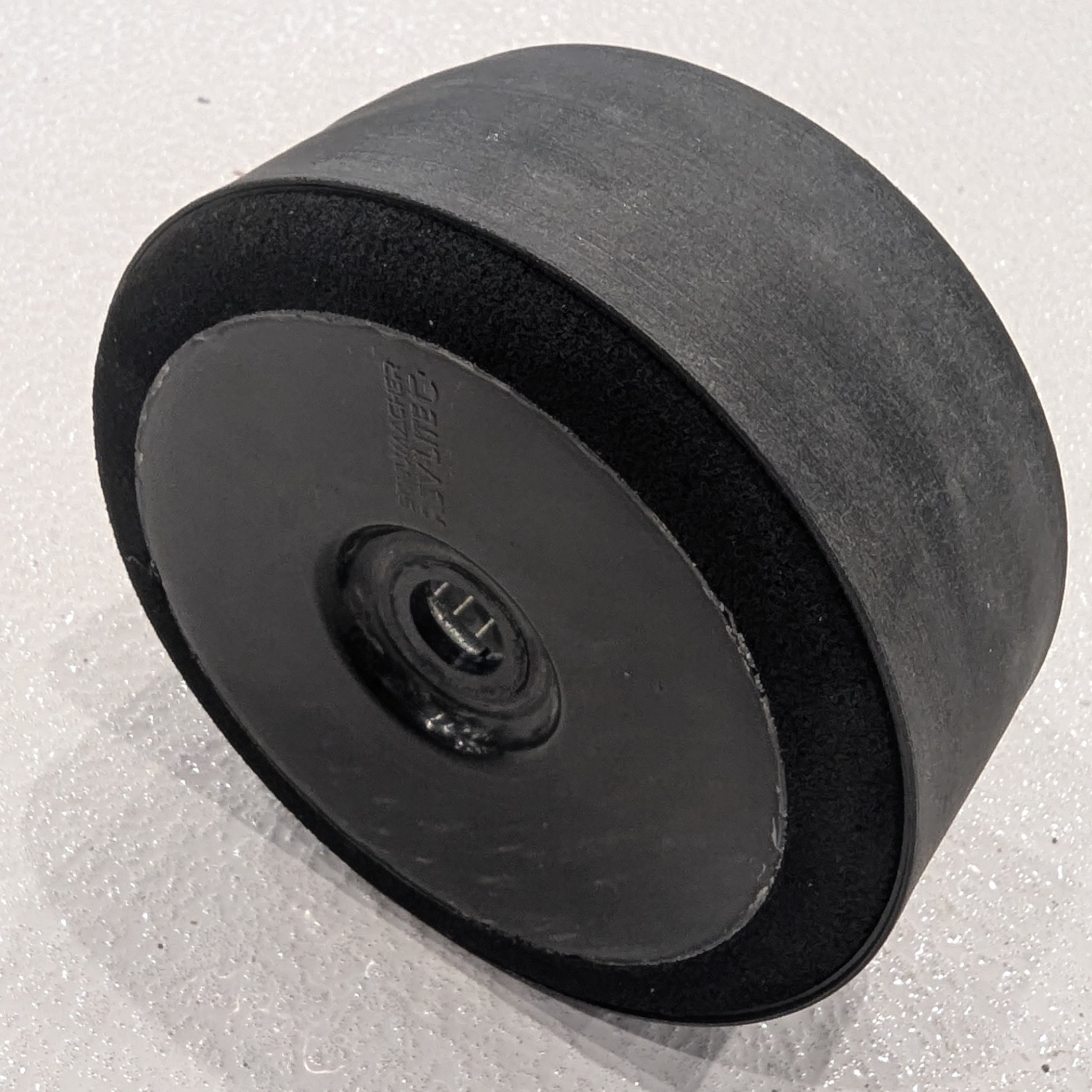}
        \caption{TPU Tire}
    \end{subfigure}
    \caption{Images of the two tire types used during the experiments.}
\end{figure}

\clearpage

\section{Controller Comparison}
\label{sec:app_base_controller}
% \begin{figure}[ht]
%     \begin{center}
%         \includegraphics[width=\textwidth]{images/laptime_comp.png}
%     \end{center}
%     \caption{Training curves for different baseline controllers, with respective reference for best lap time for the baseline controller on its own. Only lap times without boundary violations are recorded.}
%     \label{fig:ctrl_comp2}
% \end{figure}
\begin{table}[h]
    \centering
    \begin{tabular}{lccccccc}
        \toprule
        Controller & $t_{\text{min}}$ [s] ↓ & $t_{\text{max}}$ [s] ↓ & $t_{\mu}$ [s] ↓ & $\sigma$ [s] ↓ & $\text{n}_\text{bound}$ ↓ & T [min] \\
        \midrule
        MAP Baseline & 10.41 & 10.71 & 10.50 & \textbf{0.060} & \textbf{0} & --- \\
        RL MAP & \textbf{9.26} & \textbf{10.07} & \textbf{9.56} & 0.180 & \textbf{0} & 20.4 \\
        \midrule
        PP Baseline & 12.06 & 12.18 & 12.13 & \textbf{0.033} & \textbf{0} & --- \\
        RL PP & \textbf{9.81} & \textbf{11.21} & \textbf{10.46} & 0.385 & 4 & 28.2 \\
        \midrule
        FTG Baseline & 13.58 & 14.96 & 14.37 & 0.584 & \textbf{1} & --- \\
        RL FTG & \textbf{10.88} & \textbf{12.03} & \textbf{11.46} & \textbf{0.321} & 7 & 34.8 \\
        \midrule
        E2E RL & 12.04 & 13.16 & 12.33 & 0.261 & \textbf{0} & 28.8\\
        E2E RL & \textbf{9.65} & \textbf{10.36} & \textbf{9.97} & \textbf{0.183} & 4 & 81.7 \\
        \bottomrule
    \end{tabular}
    \caption{Performance comparison of different baseline residual-controller combinations as well as the \gls{e2e} controller on the \texttt{C-track} with Turbo tires after one battery life of training during 20 boundary-violation-free laps. $t_{\text{min}}$ represents the minimum lap time, $t_{\text{max}}$ the maximum lap time, $t_{\mu}$ the average lap time, $\sigma$ the standard deviation of the lap times, $n_{\text{bound}}$ the number of boundary violations until 20 violation-free laps are reached, and $T$ the total training time.}
    \label{tab:baseline_comparison}
\end{table}

\clearpage
\section{End-to-end Extended Results}
\label{app:e2e}
\begin{figure}[htbp]
    \centering
    % First figure
    \begin{subfigure}{0.7\textwidth}
        \includegraphics[width=\linewidth]{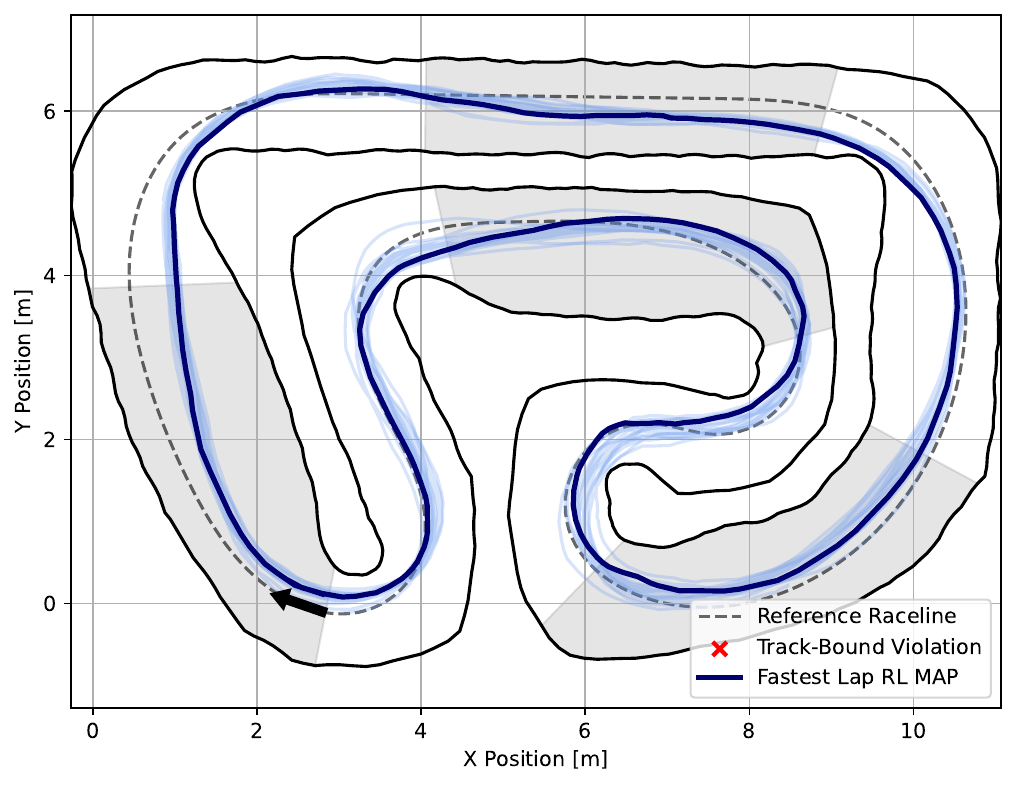}
    \end{subfigure}
    \hfill
    % Second figure
    \begin{subfigure}{0.7\textwidth}
        \includegraphics[width=\linewidth]{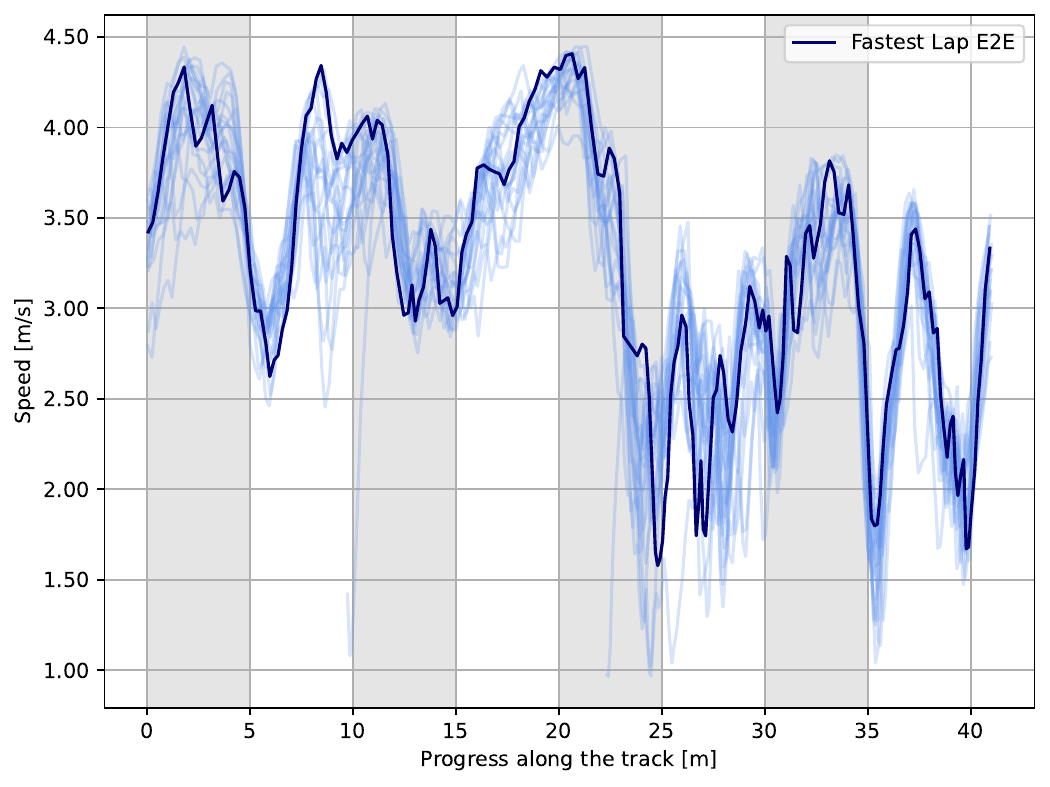}
    \end{subfigure}
    \caption{Visualizations of the trajectory and speed profile of the \gls{e2e} agent after 29\,min of training on the \texttt{C-track} with Turbo tires during deployment.}
\end{figure}
\clearpage

%% DEPLOY 84 min
\begin{figure}[H]
    \centering
    % First figure
    \begin{subfigure}{0.7\textwidth}
        \includegraphics[width=\linewidth]{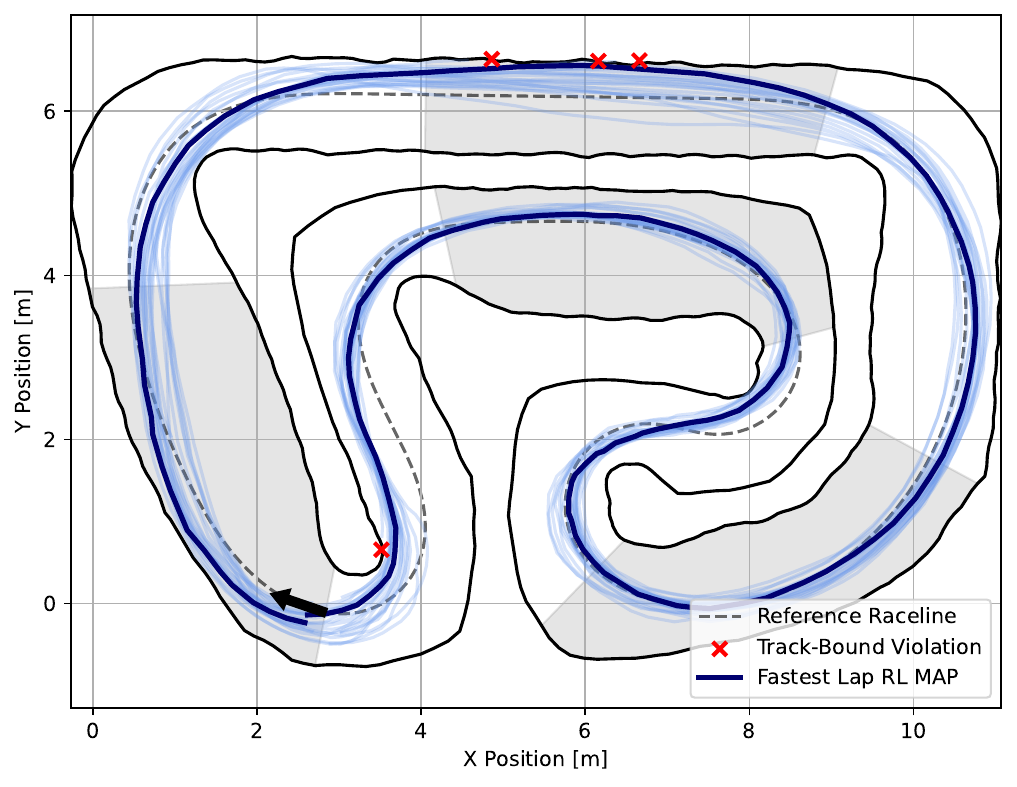}
    \end{subfigure}
    \hfill
    % Second figure
    \begin{subfigure}{0.7\textwidth}
        \includegraphics[width=\linewidth]{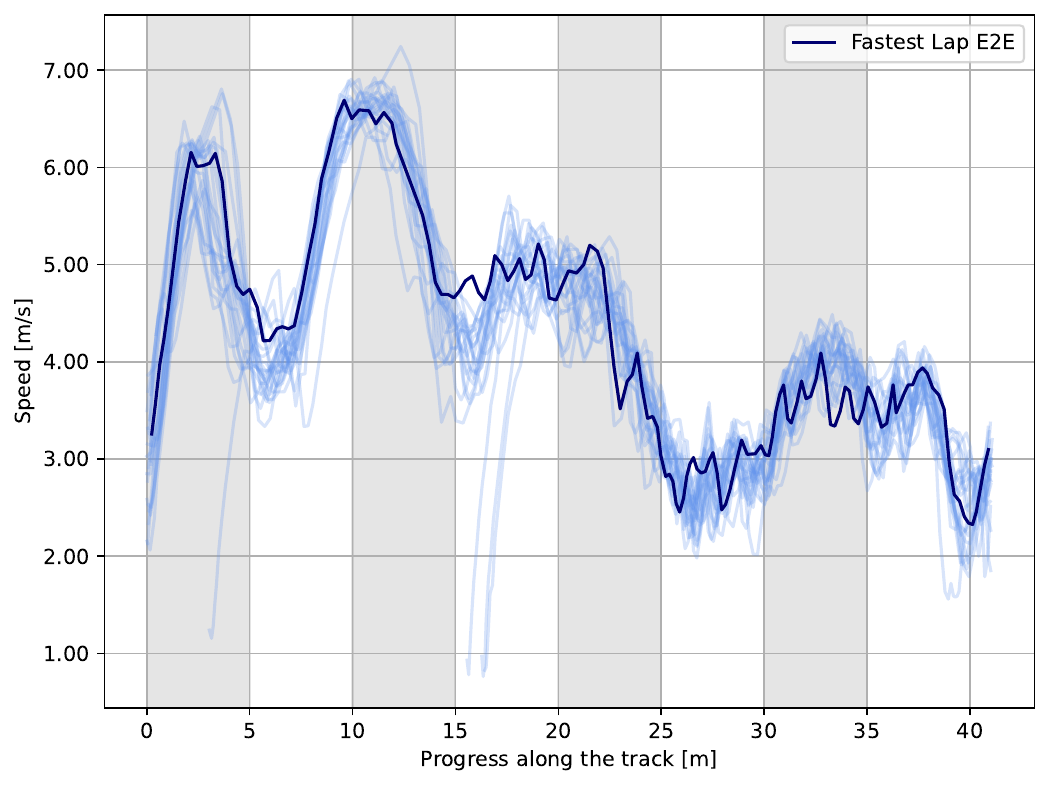}
    \end{subfigure}
    \caption{Visualizations of the trajectory and speed profile of the \gls{e2e} agent after 82\,min of training on the \texttt{C-track} with Turbo tires during deployment.}
\end{figure}

\begin{figure}[H]
    \begin{center}
        \includegraphics[width=0.6\textwidth]{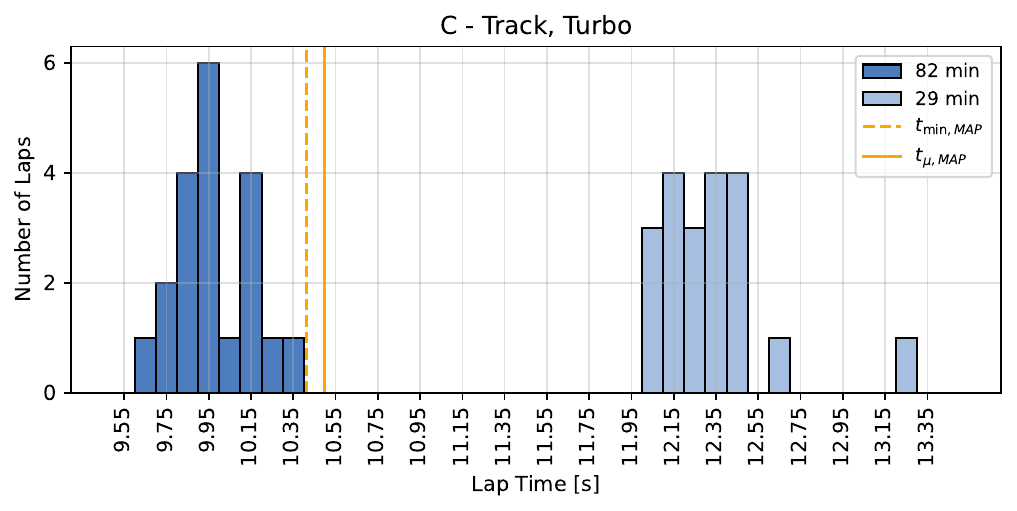}
    \end{center}
     \caption{Histogram of lap times of the \gls{e2e} controller after \qty{29}{\minute} and \qty{82}{\minute} . Orange lines represent the \gls{map} baseline (Min. Lap Time: dashed, Avg. Lap Time: solid).}
\end{figure}

\clearpage

\section{\gls{rl} \gls{map} Evaluation - Extended Results}
\label{sec:app_scratch}

\begin{table}[h]
    \centering
    \begin{tabular}{lcccccc}
        \toprule
        Run & $t_\text{min}$ [s] ↓ & $t_\text{max}$ [s] ↓ & $t_\mu$ [s] ↓ & $\sigma$ [s] ↓ & $\text{n}_\text{bound}$ ↓ & T [min] \\
        \midrule
        \textbf{C-Track, Turbo tires} \\
        \gls{map} Baseline & 10.41 & 10.71 & 10.50 & \textbf{0.060} & \textbf{0} & --- \\
        Seed 1 & 9.46 & 10.07 & 9.67 & 0.184 & \textbf{0} & 22.1 \\
        Seed 2 & \textbf{9.26} & 9.89 & 9.54 & 0.183 & \textbf{0} & 20.4 \\
        Seed 3 & 9.28 & \textbf{9.67} & \textbf{9.47} & 0.111 & 2 & 20.4 \\
        Combined Depl. & 9.26 & 10.07 & 9.56 & 0.181 & 2 & 21.0 \\
        \midrule
        \textbf{C-Track, TPU tires} \\
        \gls{map} Baseline & 12.93 & 13.78 & 13.45 & 0.338 & \textbf{0} & --- \\
        Seed 1 & 12.26 & \textbf{12.59} & \textbf{12.42} & \textbf{0.084} & \textbf{0} & 32.7 \\
        Seed 2 & 12.36 & 12.77 & 12.55 & 0.133 & \textbf{0} & 33.4 \\
        Seed 3 & \textbf{12.23} & 12.96 & 12.54 & 0.215 & 1 & 33.2 \\
        Combined Depl. & 12.23 & 12.96 & 12.50 & 0.163 & 1 & 33.1 \\
        \midrule
        \textbf{Y-Track, Turbo Tires} \\
        \gls{map} Baseline & 8.55 & 8.68 & 8.62 & \textbf{0.045} & \textbf{0} & --- \\
        Seed 1 & 7.57 & 7.96 & \textbf{7.76} & 0.126 & 1 & 24.8 \\
        Seed 2 & \textbf{7.87} & \textbf{8.26} & 8.05 & 0.126 & \textbf{0} & 21.7 \\
        Seed 3 & 7.87 & 8.46 & 8.11 & 0.151 & 5 & 21.2 \\
        Combined Depl. & 7.57 & 8.46 & 7.97 & 0.205 & 6 & 22.6\\
        \bottomrule
    \end{tabular}
    \caption{Performance of \gls{rl} \gls{map} across different tire-track combinations and weight initializations. $t_{\text{min}}$ represents the minimum lap time, $t_{\text{max}}$ the maximum lap time, $t_{\mu}$ the average lap time, $\sigma$ the standard deviation of the lap times, $n_{\text{bound}}$ the number of boundary violations until 20 violation-free laps are reached, and $T$ the total training time.}
    \label{tab:data_scratch}
\end{table}

%       MPC Baseline & 13.14 & 13.51 & 13.25 & 0.179 & \textbf{0} & --- \\

\begin{figure}[ht]
    \begin{center}
        \includegraphics[width=\textwidth]{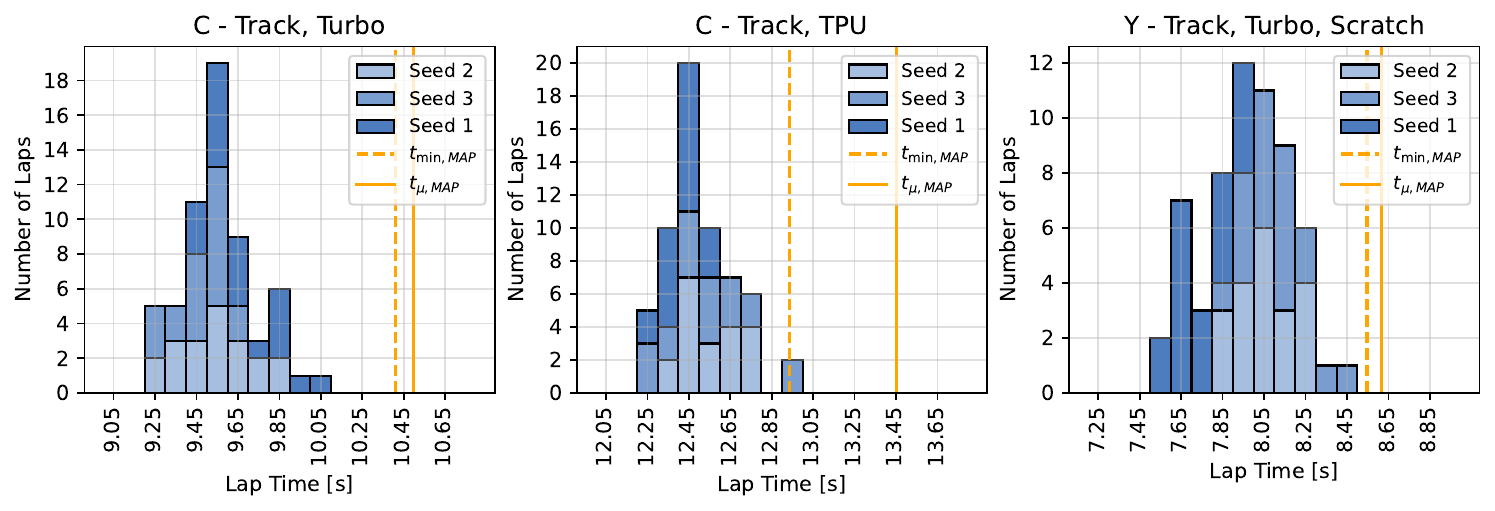}
    \end{center}
    \caption{Stacked histogram of lap times across the different tire-track combinations and weight initializations, corresponding to Table \ref{tab:data_scratch}. The orange lines represent the \gls{map} baseline.}
    \label{fig:hist_scratch}
\end{figure}

\begin{figure}[htbp]
    \centering
    % First figure
    \begin{subfigure}{\textwidth}
        \caption{C-Track, Turbo tires}
        \includegraphics[width=\linewidth]{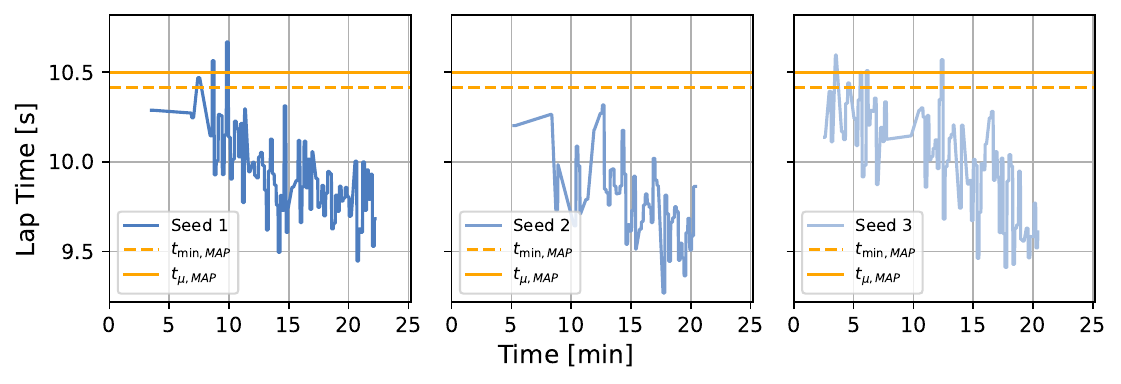}
    \end{subfigure}
    \hfill
    \newline
    % Second figure
    \begin{subfigure}{\textwidth}
        \caption{C-Track, TPU tires}
        \includegraphics[width=\linewidth]{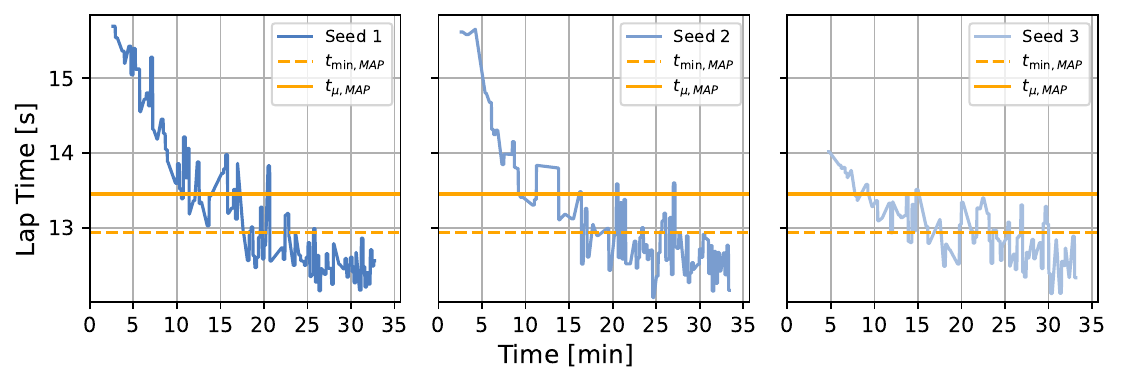}
    \end{subfigure}
    \hfill
    \newline
    % Third figure
    \begin{subfigure}{\textwidth}
        \caption{Y-Track, Turbo tires}
        \includegraphics[width=\linewidth]{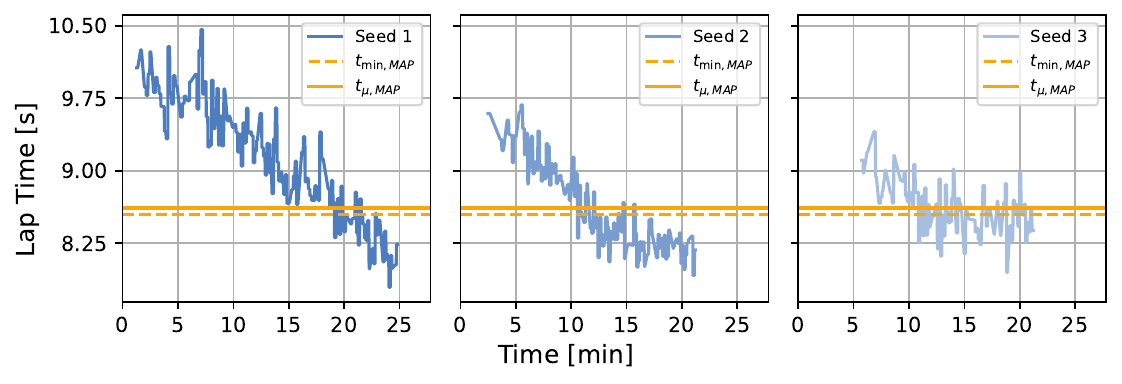}
    \end{subfigure}
    \caption{Lap Time vs. Wall Time during training for \gls{rl} \gls{map} across different track-tire combination and seed initialization, corresponding to table \ref{tab:data_scratch}. The orange lines represent the \gls{map} baseline.}
    \label{fig:laptimevstime_training}
\end{figure}

\begin{figure}[htbp]
    \centering
    % First figure
    \begin{subfigure}{0.7\textwidth}
        \includegraphics[width=\linewidth]{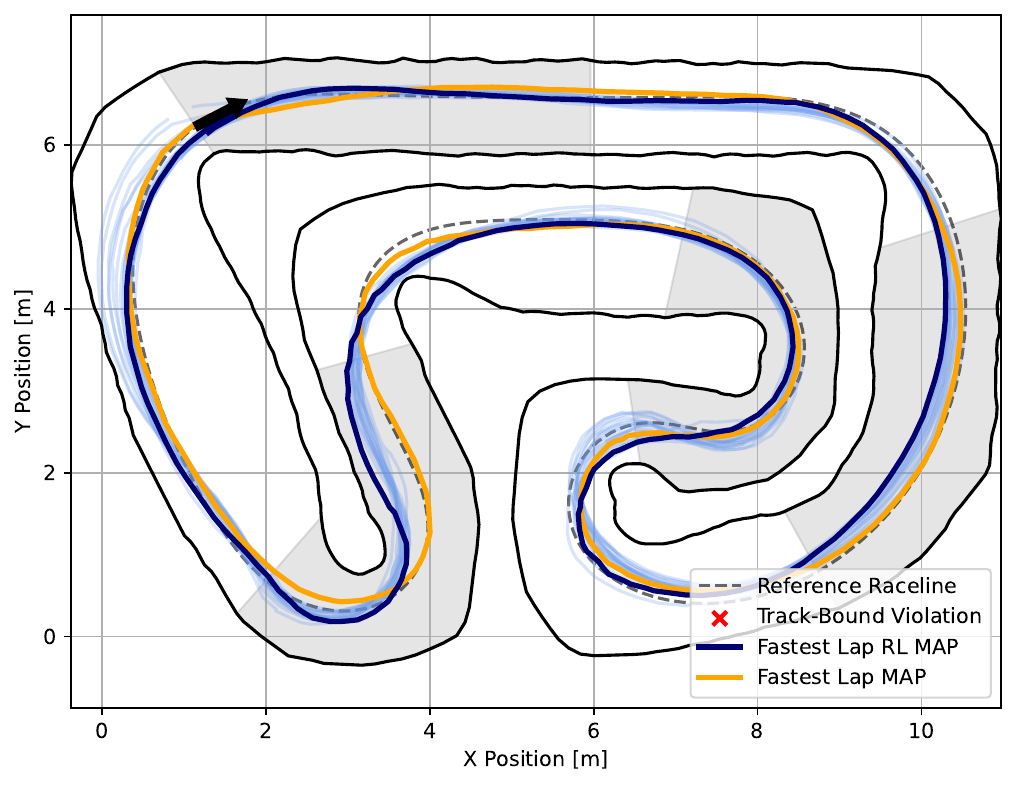}
    \end{subfigure}
    \hfill
    % Second figure
    \begin{subfigure}{0.7\textwidth}
        \includegraphics[width=\linewidth]{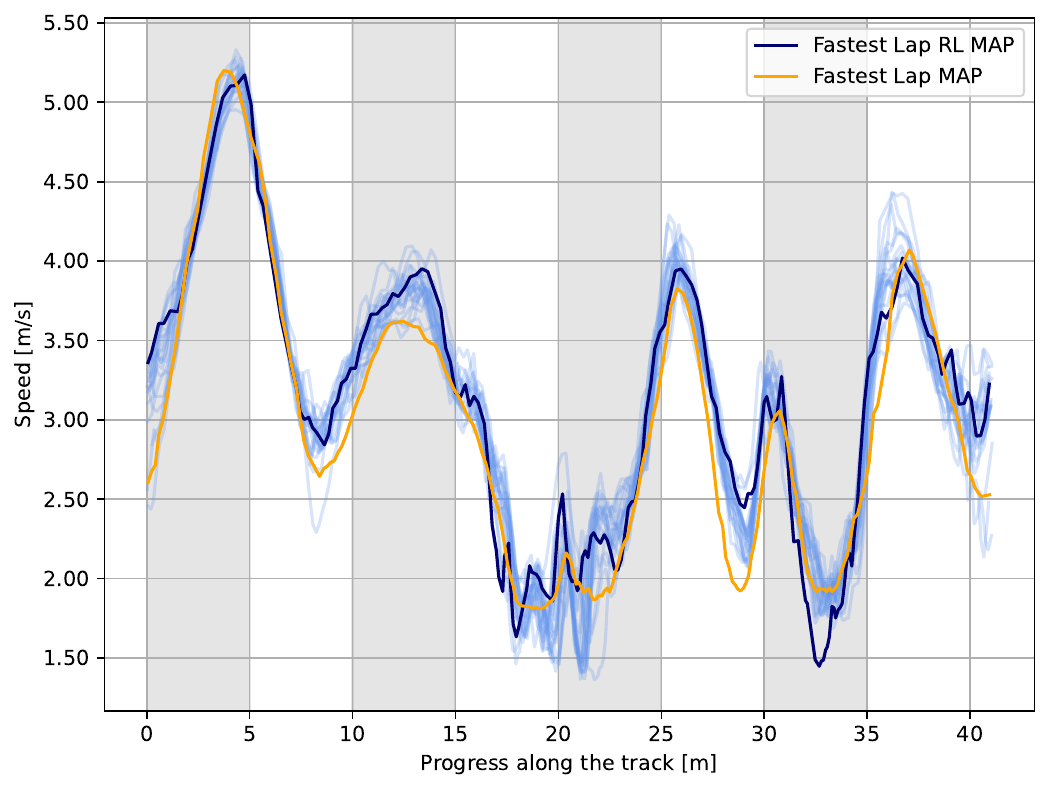}
    \end{subfigure}
    % Third figure
    \begin{subfigure}{0.7\textwidth}
        \includegraphics[width=\linewidth]{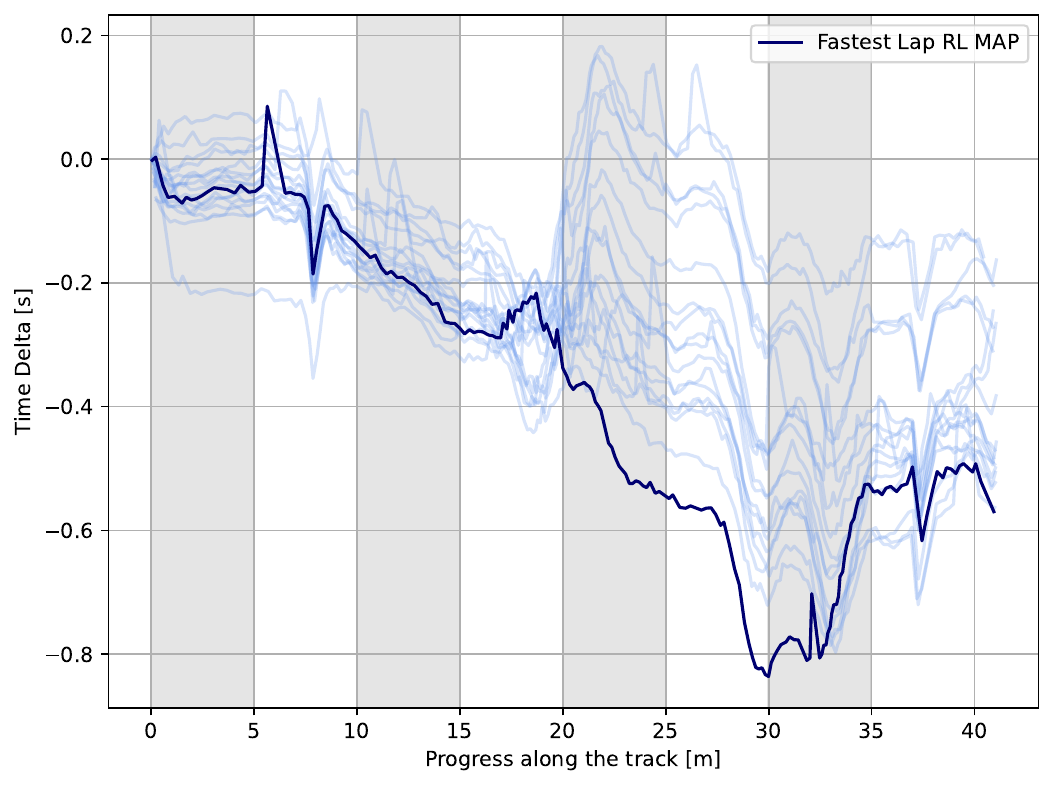}
    \end{subfigure}
    \caption{Visualizations of the trajectory, speed profile and time delta of \gls{rl} \gls{map} (Seed 2) after 20~minutes of training on the \texttt{C-track} with TPU tires.}
\end{figure}

%%% IPZ TURBO
\begin{figure}[htbp]
    \centering
    % First figure
    \begin{subfigure}{0.7\textwidth}
        \includegraphics[width=\linewidth]{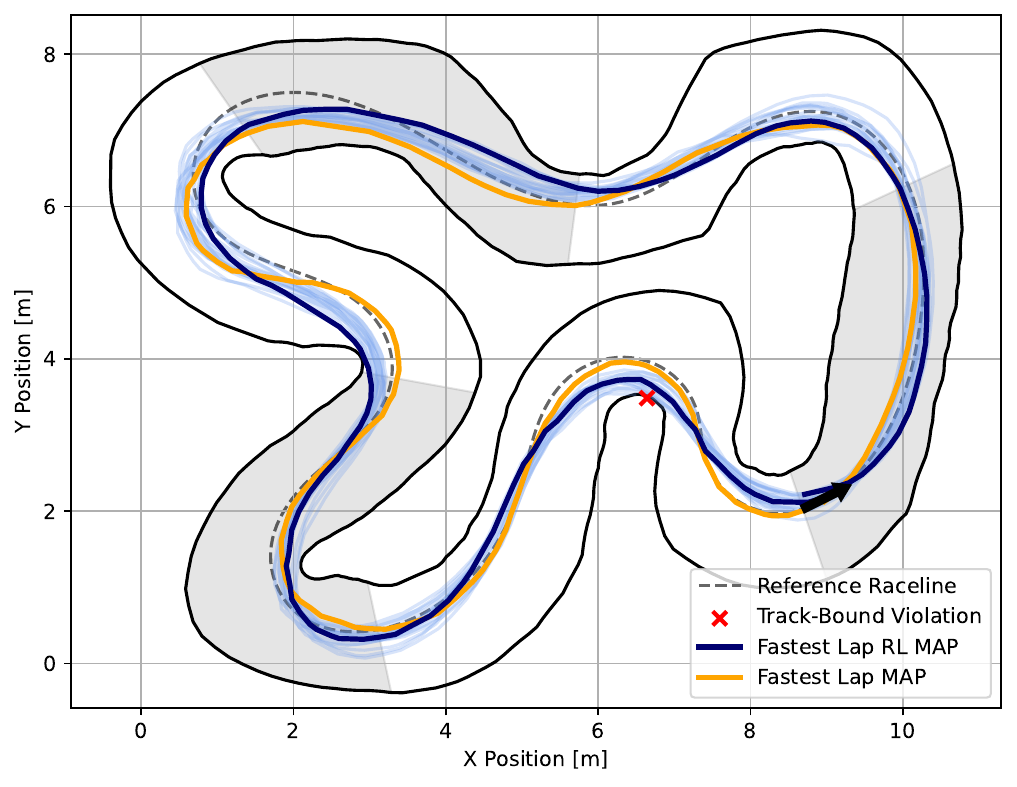}
    \end{subfigure}
    \hfill
    % Second figure
    \begin{subfigure}{0.7\textwidth}
        \includegraphics[width=\linewidth]{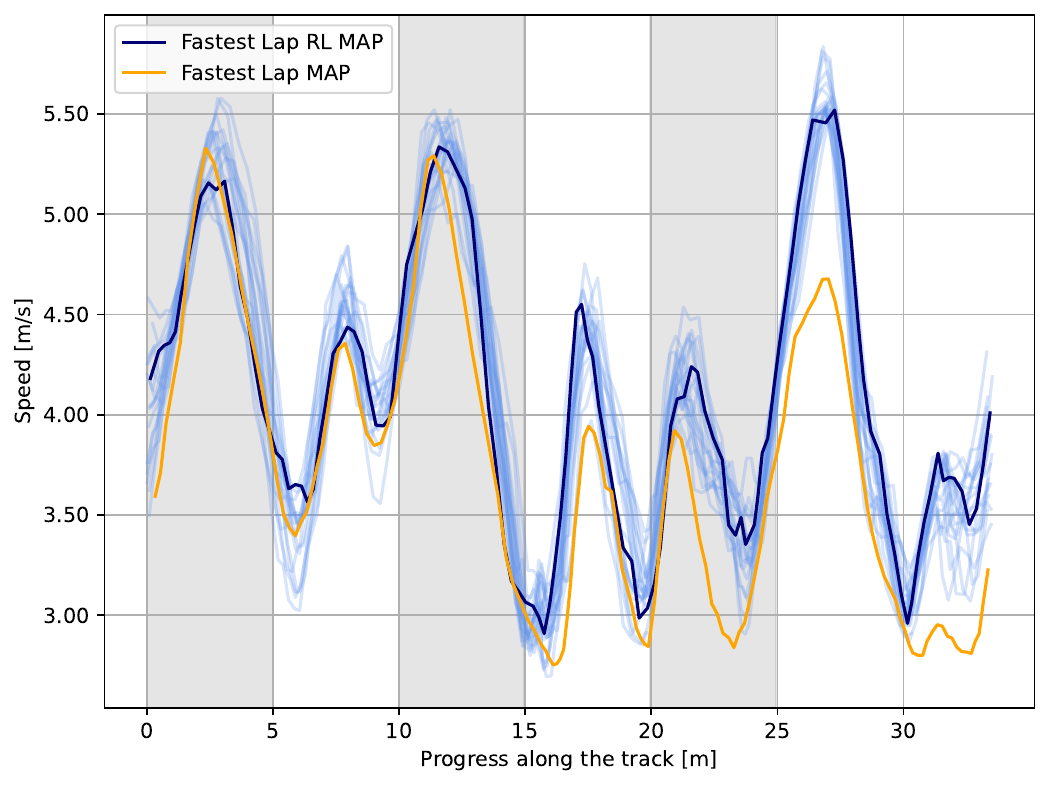}
    \end{subfigure}
    \hfill
    % Third figure
    \begin{subfigure}{0.7\textwidth}
        \includegraphics[width=\linewidth]{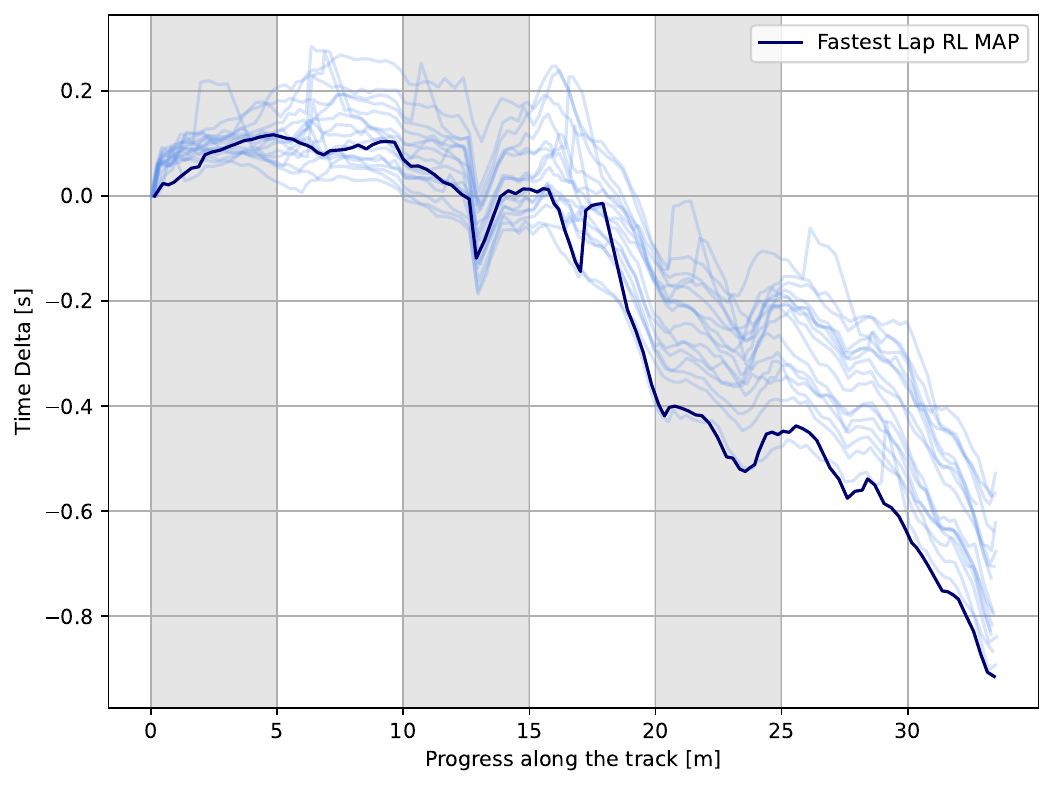}
    \end{subfigure}
    \label{fig:rbring_tpu_scratch_deploy_1_part1}
    \caption{Visualizations of the trajectory, speed profile and time delta of \gls{rl} \gls{map} (Seed 1) after 20~minutes of training on the \texttt{Y-track} with Turbo tires.}
\end{figure}
\clearpage

\section{Generalization}
\label{sec:app_generalization}

Extended results, reporting every seed's performance statistic on the generalization experiments are reported in \Cref{tab:data_ipz}.

\textbf{Learning from scratch} refers to the default training conducted in the rest of this work, and as such is a repetition of the results in \Cref{tab:data_scratch}, reported for ease of comparison. 
\textbf{Zero-Shot Transfer} refers to a policy trained on \texttt{C-track} and then deployed on \texttt{Y-track} without any additional training step, while \textbf{Few-Shot Transfer} refers to a policy trained on \texttt{C-track} and then deployed on \texttt{Y-track} after 20 minutes of additional training.

\begin{table}[ht]
    \centering
    \begin{tabular}{lcccccc}
        \toprule
        Run & $t_\text{min}$ [s] ↓ & $t_\text{max}$ [s] ↓ & $t_\mu$ [s] ↓ & $\sigma$ [s] ↓ & $\text{n}_\text{bound}$ ↓ & T [min] \\
        \midrule
        MAP Baseline & 8.55 & 8.68 & 8.62 & \textbf{0.045} & \textbf{0} & --- \\
        \midrule
        \textbf{Learning from scratch} \\
        Seed 1 & \textbf{7.57} & \textbf{7.96} & 7.76 & 0.126 & 1 & 24.8 \\
        Seed 2 & 7.87 & 8.26 & 8.05 & 0.126 & \textbf{0} & 21.6 \\
        Seed 3 & 7.87 & 8.46 & 8.11 & 0.151 & 5 & 21.2 \\
        Combined Depl. & 7.57 & 8.46 & 7.97 & 0.205 & 6 & 22.6 \\
        \midrule
        \textbf{Zero-Shot Transfer} \\
        Seed 1 & 7.99 & 8.68 & 8.37 & 0.197 & \textbf{1} & --- \\
        Seed 2 & \textbf{7.59} & \textbf{7.94} & \textbf{7.78} & 0.112 & 14 & --- \\
        Seed 3 & 8.21 & 8.58 & 8.41 & 0.189 & 7 & --- \\
        Combined Depl. & 7.59 & 8.68 & 8.10 & 0.339 & 22 & --- \\
        \midrule
        \textbf{Few-Shot Transfer} \\
        Seed 1 & \textbf{7.63} & 8.13 & 7.85 & 0.138 & 2 & 17.7 \\
        Seed 2 & 7.73 & 8.20 & \textbf{7.96} & 0.114 & \textbf{0} & 21.6 \\
        Seed 3 & 7.81 & \textbf{8.12} & 7.99 & 0.088 & 2 & 19.7 \\
        Combined Depl. & 7.63 & 8.20 & 7.93 & 0.129 & 4 & 19.6 \\
        \bottomrule
    \end{tabular}
    \caption{Performance of \gls{rl} \gls{map} with few-shot and zero-shot transfer from the \texttt{C-track} to the \texttt{Y-track}. $t_{\text{min}}$ represents the minimum lap time, $t_{\text{max}}$ the maximum lap time, $t_{\mu}$ the average lap time, $\sigma$ the standard deviation of the lap times, $n_{\text{bound}}$ the number of boundary violations until 20 violation-free laps are reached, and $T$ the total training time.}
    \label{tab:data_ipz}
\end{table}

\begin{figure}[ht]
    \centering
    % First figure
    \begin{subfigure}{0.6\textwidth}
        \includegraphics[width=\linewidth]{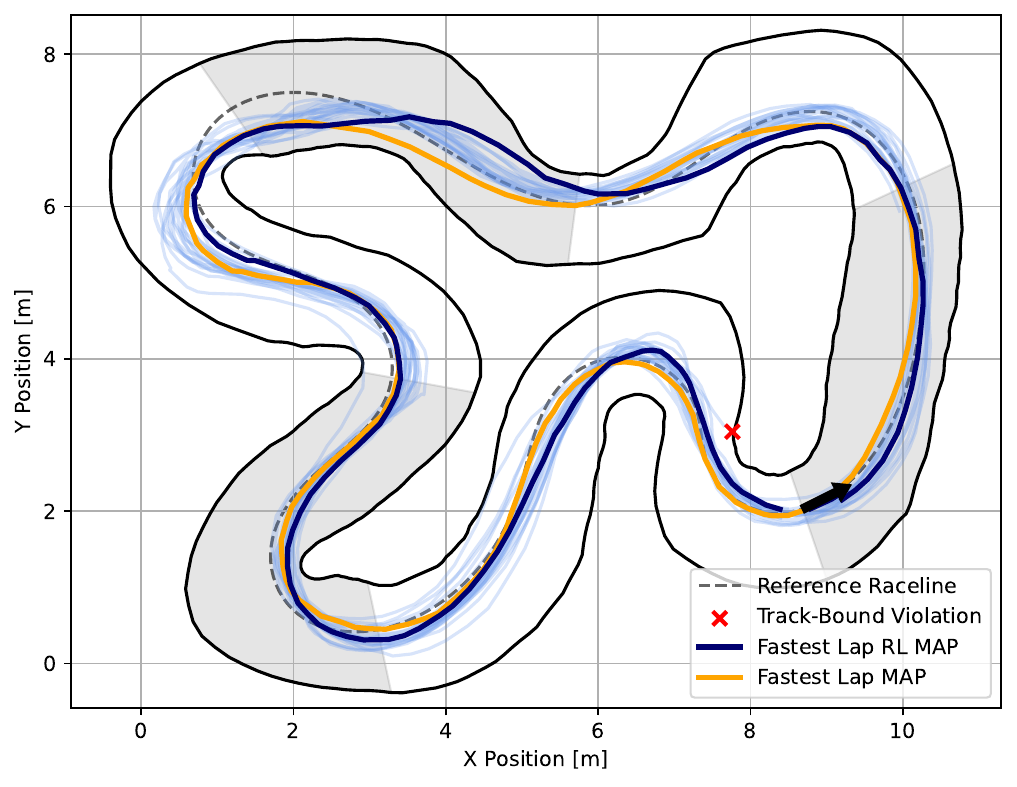}
    \end{subfigure}
    \hfill
    % Second figure
    \begin{subfigure}{0.6\textwidth}
        \includegraphics[width=\linewidth]{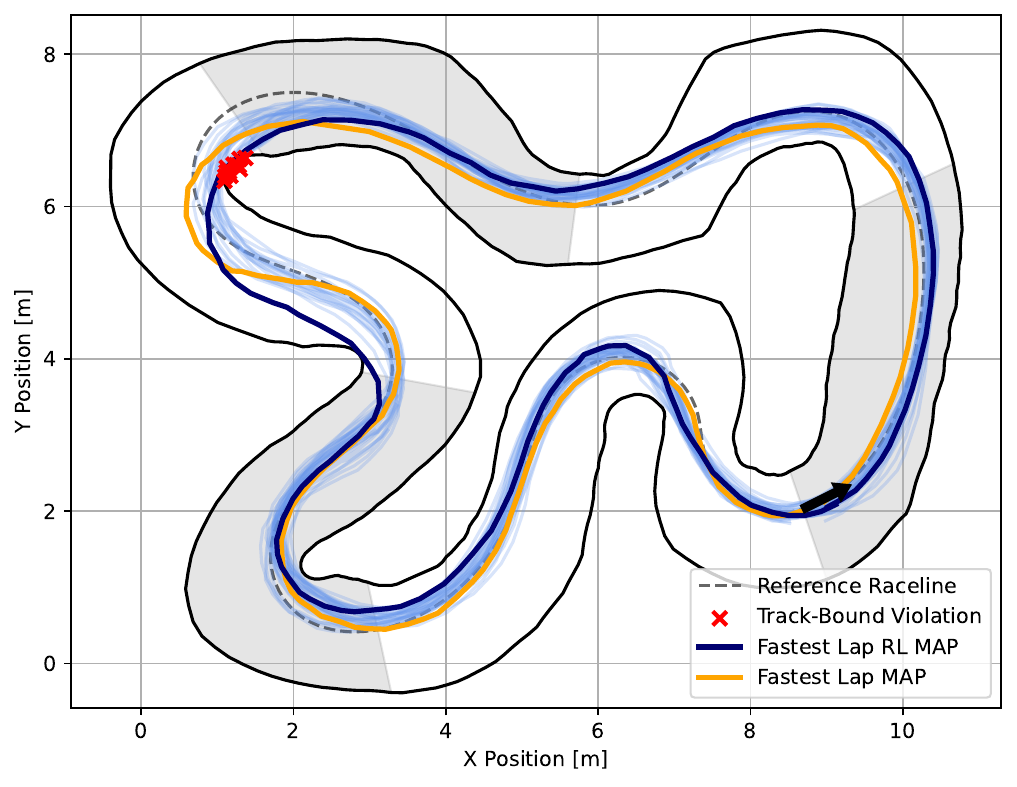}
    \end{subfigure}
    \hfill
    % Third figure
    \begin{subfigure}{0.6\textwidth}
        \includegraphics[width=\linewidth]{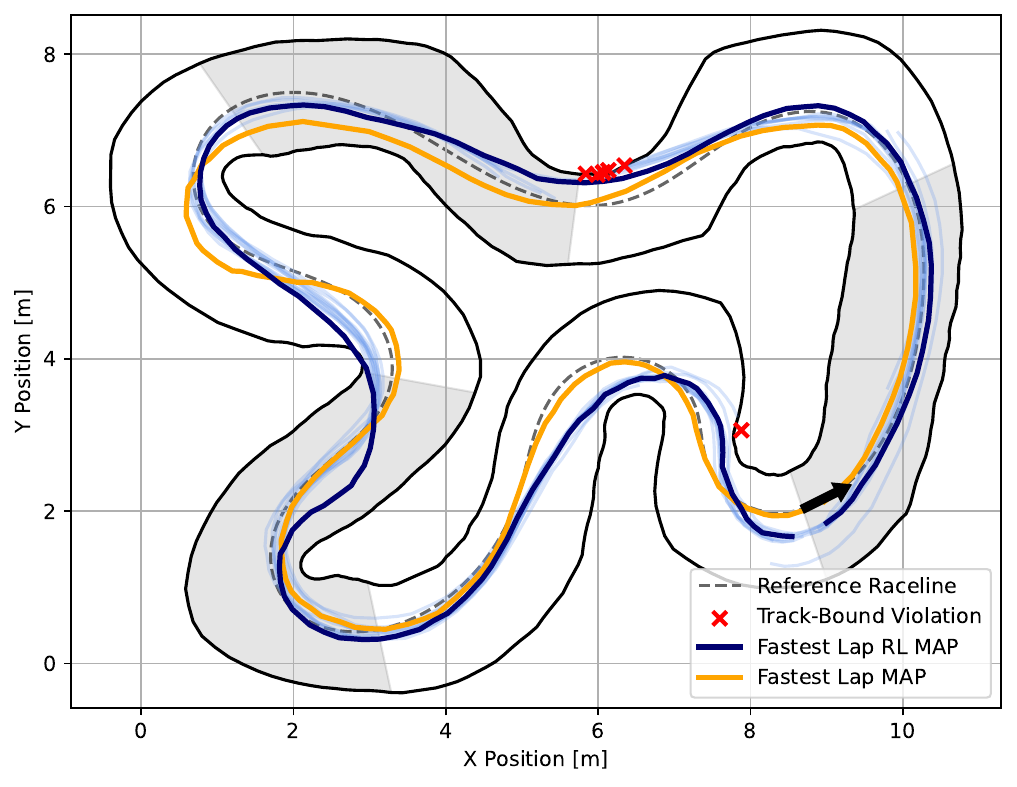}
    \end{subfigure}
    \caption{Visualizations of the trajectories during zero-shot transfer from \texttt{C-track} to \texttt{Y-track}  for Seeds 1, 2 and 3 (from top to bottom).}
    \label{fig:rbring_tpu_zero_deploy_1_part1}
\end{figure}

\clearpage

\section{Heuristic Delayed Reward Adjustment}
\label{appendix:dra}
\begin{algorithm}[h]
    \caption{Heuristic Delayed Reward Adjustment}
    \label{alg:reward_adjustment}
    \begin{algorithmic}[1]
        \Require Replay buffer \texttt{replay\_buffer}, current step index \texttt{current\_step}
        \State \textbf{Parameters:} \texttt{adjustment\_steps} $N$, \texttt{penalty} $\text{p}$
        \If{\texttt{terminal\_state} \textbf{is} \texttt{True} \textbf{and} \texttt{replay\_buffer.reward[current\_step]} $=$ \texttt{0}}
            \For{$n = 0$ \textbf{to} $N - 1$}
                \State \texttt{previous\_step} $\gets$ \texttt{current\_step} $-\;n$
                \State \texttt{current\_reward} $\gets$ \texttt{replay\_buffer.reward[current\_step]}
                \State \texttt{decaying\_penalty} $\gets$ $\text{p} - n(\text{p}/N) $
                \State \texttt{adjusted\_reward} $\gets$ \texttt{current\_reward} $-$ \texttt{decaying\_penalty}
                \State \texttt{replay\_buffer.reward[current\_step} $-\;n$\texttt{]} $\gets$ \texttt{adjusted\_reward}
            \EndFor
        \EndIf
    \end{algorithmic}
\end{algorithm}

% \begin{algorithm}[h]
%     \caption{Heuristic Delayed Reward Adjustment}
%     \label{alg:reward_adjustment}
%     \begin{algorithmic}[1]
%         \Require Replay buffer \texttt{replay\_buffer}, current step index \texttt{current\_step}
%         \State \textbf{Parameters:} \texttt{adjustment\_steps} $N$, \texttt{penalty} $\text{p}$
%         \If{\texttt{terminal\_state} \textbf{is} \texttt{True} \textbf{and} \texttt{replay\_buffer.reward[current\_step]} $=$ \texttt{0}}
%             \For{$n = 0$ \textbf{to} $N - 1$}
%                 \State \texttt{previous\_step} $\gets$ \texttt{current\_step} $-\;n$
%                 \State \texttt{current\_reward} $\gets$ \texttt{replay\_buffer.reward[current\_step]}
%                 \State \texttt{decaying\_penalty} $\gets$ $\text{p} - n(\text{p}/N) $
%                 \State \texttt{adjusted\_reward} $\gets$ \texttt{current\_reward} $-$ \texttt{decaying\_penalty}
%             \EndFor
%         \EndIf
%     \end{algorithmic}
% \end{algorithm}

\section{Ablation Study Extended Results}
\label{appendix:abl}
\begin{table}[h]
    \centering
    \begin{tabular}{lcccccc}
        \toprule
        Run & $t_\text{min}$ [s] ↓ & $t_\text{max}$ [s] ↓ & $t_\mu$ [s] ↓ & $\sigma$ [s] ↓ & $\text{n}_{\text{laps},\,\mu}$ & $\text{n}_{\text{bound},\,\mu}$ ↓ \\
        \midrule
        HDRA+TD3, Sync & 10.53 & 13.82 & 12.09 & 0.691 & 12.4 & 3.4 \\
        HDRA+TD3, Async & \textbf{9.12} & \textbf{10.26 }& \textbf{9.53} & \textbf{0.246} & \textbf{16.0} & \textbf{2.2} \\
        \midrule
        TD1, Async & 9.73 & 20.81 & 11.53 & 2.823 & 11.0 & 5.6 \\
        TD3, Async & \textbf{9.29} & \textbf{10.27 }& \textbf{9.68} & \textbf{0.214} & \textbf{16.2} & \textbf{2.0} \\
        \bottomrule
    \end{tabular}
    \caption{Performance metrics summarized over five independent training runs, evaluating the last three minutes of \qty{40}{\minute}-training runs during the ablation study on the simulated \texttt{C-Track}. The overall minimum, maximum, average lap time, and standard deviation are indicated with $t_\text{min},\,t_\text{max},\,t_\mu,\,\sigma$ respectively. The average over five runs of the number of laps and the number of boundary collisions in the three minutes is indicated with $\text{n}_{\text{laps},\,\mu}$ and $\text{n}_{\text{bound},\,\mu}$ respectively.}
    \label{tab:ablation}
\end{table}

        %HDRA+TD3, Sync & 10.53 & 13.82 & 12.09 & 0.691 & 62 & 17 \\
        %HDRA+TD3, Async & \textbf{9.12} & \textbf{10.26 }& \textbf{9.53} & %\textbf{0.246} & \textbf{80} & \textbf{11} \\
        %TD1, Async & 9.73 & 20.81 & 11.53 & 2.823 & 55 & 28 \\
        %TD3, Async & \textbf{9.29} & \textbf{10.27 }& \textbf{9.68} & %\textbf{0.214} & \textbf{81} & \textbf{10} \\

\clearpage

\section{Implementation Details}
\label{appendix:details}
\begin{table}[ht]
    \centering
    \begin{tabular}{llc}
        \toprule
        Category & Parameter & Value \\
        \midrule
        \textbf{Observations} & Number of reference points J & 20 \\
                                  & Look-ahead horizon l & 6 m\\
        \midrule
        \textbf{Reward Design} & Reward multiplier for progress $\lambda$ & 10 \\
                               & Penalty $\text{p}$ & 10\\
                               & \gls{hdra} steps N & 10\\
        \midrule
        \textbf{Safety Mechanisms} & Safety filter $\psi_{\text{filter}}$ & $[\pi/6;\pi/2]$ rad \\
                                   & Increment/decrement rate $\epsilon$ & 0.05 rad \\
        \midrule
        \textbf{SAC Parameters} & Optimizer & Adam \\
                            & Learning rate & 0.003 \\
                            & Discount factor $\gamma$ & 0.96 \\
                            & Temporal-difference (TD) steps & 3 \\
                            & Replay buffer size & $1e^6$\\
                            & Batch size & 256 \\
                            & Number of hidden layers & 2\\
                            & Size of hidden layers & 256\\
                            & Activations functions & ReLU\\
        \bottomrule
    \end{tabular}
    \caption{Hyperparameters for on-board \gls{rl}.}
    \label{tab:hyperparameters}
\end{table}

\section{\gls{mpc} Implementation Details}
\label{appendix:mpc_formulation}
The \gls{mpc} formulation used in this work uses a single-track bicycle model, with tire forces modeled with the \emph{Pacejka Magic Formula} \citep{PACEJKA20121}. The cost function is derived from the \gls{mpc} formulation of \cite{vazquez2020mpccurv}.

The model, expressed in curvilinear coordinates, reads as follows:
\begin{align}
\dot s &=  ((v_x  \cos\varphi_{ref}) - (v_y  \sin(\varphi_{ref}))) / (1 - \kappa_{ref}(s)\,n) \\
\dot n &=v_x \sin\varphi_{ref}+v_y \cos \varphi_{ref} \\
\dot \varphi_{ref} &=r - \kappa_{ref}(s)\,\dot{s} \\
\dot v_x &=a+\frac{1}{m}(-F_{yf}\sin \delta+m v_y r) \\
\dot v_y &=\frac{1}{m}(F_{yr}+F_{yf}\cos \delta-m v_x r) \\
\dot r &=\frac{1}{I_z}(F_{yf}l_f\cos \delta-F_{yr}l_r) 
\end{align}

where $s$ represents the longitudinal advancement along the reference line, $n$ the perpendicular lateral displacement, $\varphi_{ref}$ the angle relative to the tangent at the point that meets the perpendicular, and $\kappa_{ref}(s)$ the curvature of the reference line at the corresponding $s$.
The longitudinal input is modeled as $a$ and the steering is $\delta$, resulting in the input $\bm{u}=(a,\,\delta)$.
Longitudinal and lateral velocities are represented by $v_x,\,v_y$, respectively, and $r$ represents the yaw rate.
These latter three states are mostly influenced by tire forces, $F_{yf},\,F_{yr}$, indicating the lateral tire force on the front and rear lumped tires, respectively.
They follow the equations from the \emph{Pacejka Magic Formula}:
\begin{align}
F_{y,i}=&\mu F_{z,i}D_i \sin  [C_i \arctan\left(B_i \alpha_i-E_i \left(B_i \alpha_i-\arctan\left(B_i \alpha_i\right) \right) \right) ]
\end{align}
where $i\in\{front,\,rear\}$ discriminates between front and rear tire. 
The slip angles $\alpha_i$ are as follows:
\begin{align}
\alpha_f&=\arctan\left(\frac{v_y+ r l_f}{v_x}\right)-\delta \\
\alpha_r&=\arctan\left(\frac{v_y- r l_r}{v_x}\right).
\end{align}
The \emph{Pacejka} parameters $(B_i,\,C_i,\,D_i,\,E_i)$ are obtained via the procedure detailed by \cite{dikici2025ontracksysid}.

The \gls{mpc} cost function, derived from \cite{vazquez2020mpccurv}, is designed in order to follow the raceline (both on the positional and the velocity profile) computed by adjusting the minimum curvature raceline presented by \cite{Heilmeier2020MinimumCar} to the F1TENTH vehicle.
The \gls{mpc} problem then approximately reads as follows:
\begin{align}
    \min_{U} \quad & l_f(\bm{x}_{N}) + \sum_{i=0}^{N-1} l(\bm{x}_{i}, \bm{u}_{i}) \\
    \quad & \bm{x}_{i+1} = f(\bm{x}_{i},u_{i}) \; \; \forall i \in \{0,...,N-1\} \nonumber \\
    & \bm{x}_{i} \in \mathcal{X} \\
    & u_{i} \in \mathcal{U} \\
    & \bm{x}_{N} \in \mathcal{X}_{f} \\
    & \bm{x}_0 = \hat{x}
\end{align}
where:
\begin{itemize}
    \item The state is $\bm{x}=[s,\,n,\,\varphi_{ref},\,v_x,\,v_y,\,r]^\top$
    \item The number of timesteps is represented by $N$
    \item The state transition equation, derived from a \emph{Runge-Kutta} discretization of the equations above, is displayed as $f(\bm{x},\,\bm{u})$
    \item $\mathcal{X}$ represents the track boundaries and state constraints sets
    \item $\mathcal{U}$ represents the input constraints set
    \item $\hat{x}$ is the measured state
    \item the cost function is $l(\bm{x}_{i}, \bm{u}_{i}) = w_{n}n+w_{\varphi_{ref}}\varphi_{ref}+w_{v_x}(v_x-v_x^{race})$, with $v_x^{race}$ being the velocity obtained from the raceline generation
    \item $l_f(\bm{x}_{i}, \bm{u}_{i}) = 10l(\bm{x}_{i}, \bm{u}_{i})$ 
\end{itemize}

Soft constraints are further integrated on state constraints, to try to ensure feasibility.
The controller then is manually tuned for lap time minimization.
The implementation is then carried out within the \emph{acados} framework \citep{Verschueren2021}.

The \gls{mpc} controller, in the end, did not achieve the fastest performance on the tires with the highest grip, and only achieved the fastest average lap time on the tires with the lowest grip.
 We attribute this performance gap to potential \gls{s2r} mismatches, such as unmodelled delays, as well as the higher signal noise inherent to the vehicle's sensor setup and embedded system constraints, which may have compromised the \gls{mpc}'s performance and induced unnecessary conservativity.
% we believe this performance gap might be due to \gls{s2r} mismatches still present, such as unmodelled delays; furthermore, the higher noise signals to which the vehicle is subject, being entirely embedded, might also have affected the performance of the \gls{mpc} and induced some unnecessary robustness. 

\end{document}